\documentclass[10pt,twocolumn,letterpaper]{article}

\usepackage{wacv}
\usepackage{times}
\usepackage{cite}
\usepackage{epsfig}
\usepackage{graphicx}
\usepackage{amsmath}
\usepackage{amssymb}
\usepackage[export]{adjustbox}

\usepackage[pagebackref=true,breaklinks=true,letterpaper=true,colorlinks,bookmarks=false]{hyperref}

\wacvfinalcopy %

\def\wacvPaperID{421} %

\ifwacvfinal\fi
\begin{document}

\title{Using a single RGB frame for real time 3D hand pose estimation in the wild}

\author{Paschalis Panteleris$^1$ \hspace{2cm} Iason Oikonomidis$^1$ \hspace{2cm} Antonis Argyros$^{1,2}$ \\
$^1$Institute of Computer Science, FORTH \\ $^2$ Computer Science Department, UOC\\
{\tt\small \{padeler,oikonom,argyros\}@ics.forth.gr}
}

\maketitle
\ifwacvfinal\thispagestyle{empty}\fi

\newcommand{\placeholderfigure}[1]{
\begin{figure}[t]
\begin{center}
\includegraphics[width=0.99\columnwidth]{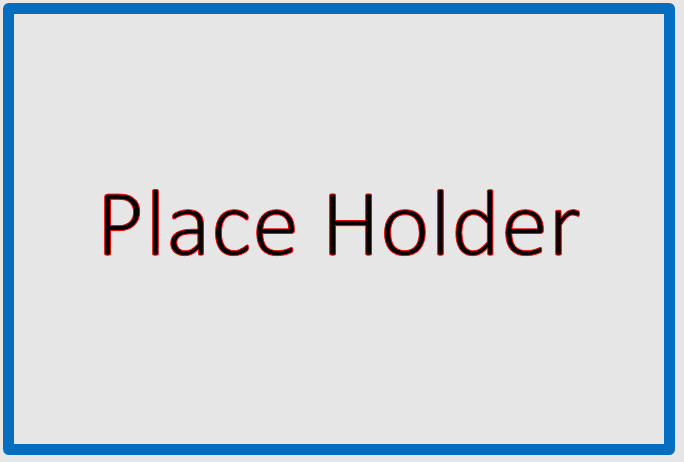}
\caption{\label{fig:#1} #1.}
\end{center}
\end{figure}
}

\newcommand{\placeholderfiguretwocols}[1]{
\begin{figure*}[t]
\begin{center}
\includegraphics[width=0.99\columnwidth]{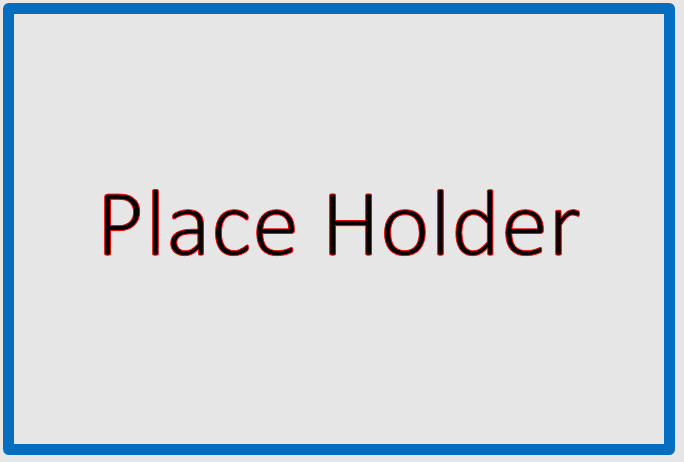}
\caption{\label{fig:#1} #1}
\end{center}
\end{figure*}
}

\begin{abstract}

We present a method for the real-time estimation of the full 3D pose of one or more human hands using a single commodity RGB camera. Recent work in the area has displayed impressive progress using RGBD input. However, since the introduction of RGBD sensors, there has been little progress for the case of monocular color input. We capitalize on the latest advancements of deep learning, combining them with the power of generative hand pose estimation techniques to achieve real-time monocular 3D hand pose estimation in unrestricted scenarios.
More specifically, given an RGB image and the relevant camera calibration information, we employ a state-of-the-art detector to localize hands. Given a crop of a hand in the image, we run the pretrained network of OpenPose for hands to estimate the 2D location of hand joints. Finally,  non-linear least-squares minimization fits a 3D model of the hand to the estimated 2D joint positions, recovering the 3D hand pose. Extensive experimental results provide comparison to the state of the art as well as qualitative assessment of the method in the wild.

\end{abstract}

\section{Introduction}

Hand pose estimation using  markerless visual input is a long-standing problem in the area of articulated object pose estimation. The first efforts to tackle the problem date as early as 1994~\cite{rehg1994visual}. The popularity of the problem as well as new insights and progress towards solving it are steadily increasing in recent years. This interest can be attributed to several factors. Firstly, the introduction of inexpensive RGBD cameras has enabled easy access to relatively high quality depth maps, facilitating the solution of an inherently ill-posed problem. The advent of Augmented Reality and Virtual Reality applications has served as a source of demand for natural user interfaces, fueling the interest in solutions to the problem of hand pose estimation. Finally, the success of deep learning approaches has enabled new levels of accuracy and robustness and lower execution times. Still, despite the significant progress that has been achieved, the problem remains unsolved in its full generality.

\begin{figure}[t]
\begin{center}
\includegraphics[width=0.99\columnwidth,frame]{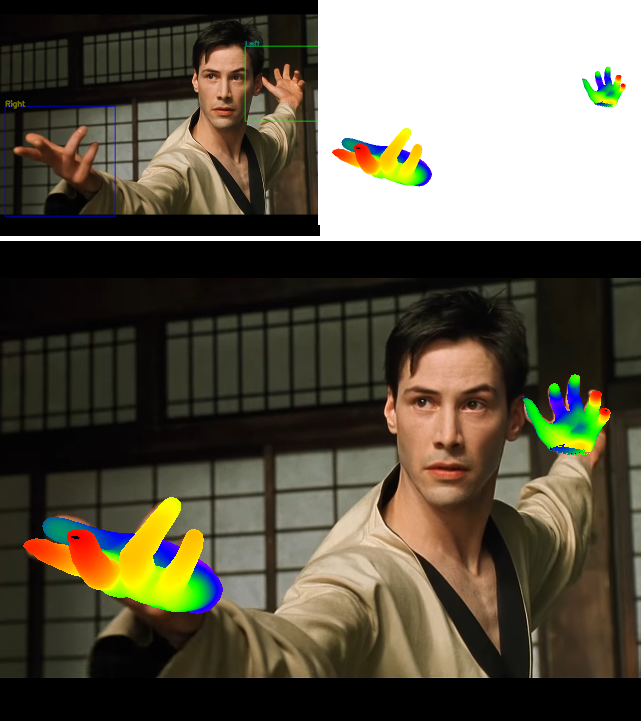}
\caption{\label{fig:overview} We propose a method that, given a single RGB image frame (top left), can estimate the 3D position and articulation of the imaged  hands (top right). The bottom image shows the output of the proposed method superimposed on the input.}
\end{center}
\end{figure}

This work tackles the problem of recovering, in real-time, the full 3D state of one or more hands, observed from a conventional RGB camera. Specifically, for each hand observed by a calibrated monocular color camera, we seek to estimate the 3D position of $21$ pre-defined key points: the base of the palm, the centers of the $15$ finger joints and the $5$ fingertips. The goal is to estimate these points in real-time or interactive frame rates, allowing the continuous re-estimation of these positions in a stream of images.
When performing this task without explicit assumptions on the motion of the hands, the approach is commonly referred to as tracking-by-detection~\cite{andriluka2008people}. The advantage of such approaches is that they do not require initialization of the tracking process. On the other hand, if the application context permits valid assumptions on the type of motion (e.g., temporal continuity, hands involved in specific activities and motions, etc), then the problem can be treated as a tracking one, facilitating the search in the large space of possible solutions.

There are several interacting factors that make this problem hard to solve~\cite{erol2007vision}. The human hand is very dexterous, and so the problem involves the estimation of more than $20$ finger articulation parameters. If the hand moves fast, purely tracking-based methods are prone to track loss. The uniform/textureless appearance of a hand hinders the identification of different parts. Additionally, the hand appearance can vary widely depending on the viewpoint of observation because of articulation, self occlusions, or occlusions from other objects during manipulation. 

To deal with these difficulties, different classes of approaches have been proposed. A common categorization (adopted for example in \cite{sridhar2013interactive}) of methods regards the type of runtime computations: methods that try to fit observed features to synthesized, hypothesized ones are called generative. Methods that pre-compute a mapping from the input data directly to the pose space, typically in a large training step, are called discriminative. Finally, methods that in some way incorporate characteristics of both of these approaches are called hybrid.

In this work we develop a hybrid approach. More specifically, we advocate the use of discriminative, state-of-the-art deep learning networks to solve the problems of 2D hand detection and 2D hand joint localization. On top of the results of these components, we employ generative model fitting that is formulated as a non-linear least-squares optimization problem and solved using the Levenberg-Marquardt optimizer. This approach enables real-time and robust tracking of the full 3D hand pose using conventional RGB input. Experimental results on ground-truth-annotated datasets as well as in youtube videos acquired in the wild show that the proposed approach outperforms state of the art solutions in accuracy and can be used for effective 3D hand pose estimation in real world situations.

\section{Related work}
The problem of hand pose estimation using visual markerless input is a long standing one in the relevant literature, exhibiting both theoretical interest and important applications. Early works \cite{rehg1994visual} required specialized hardware to achieve real-time performance. Since then, both the available computational power as well as the input modalities have significantly evolved.
The first important change happened in 2010 with the introduction of commodity RGBD sensors. The computer vision community has heavily capitalized upon this relatively inexpensive and reliable source of information to improve the accuracy and performance of several problems including body pose and hand pose estimation \cite{shotton2013real, oikonomidis2011efficient, tang2013real}.
Almost simultaneously with the advent of depth cameras, modern commodity hardware such as GPUs have enabled the practical use of deep learning, yielding a spectacular increase in the accuracy of tasks such as image classification \cite{simonyan2014very}.
Thus, extensive efforts have been devoted to the problem of estimating the 3D pose of a single hand observed using a depth sensor, possibly with the aid of a similar-viewpoint color image \cite{
oikonomidis2011efficient,
keskin2012hand,
kyriazis2013physically,
melax2013dynamics,
sridhar2013interactive,
tang2013real,
xu2013efficient,
qian2014realtime,
tang2014latent,
taylor2014user,
tompson2014real,
fleishman2015icpik,
khamis2015learning,
li20153d,
MakrisArgyros2015a,
MakrisKyriazisArgyros2015a,
oberweger2015hands,
oberweger2015training,
poier2015hybrid,
rogez2015first,
sharp2015accurate,
sridhar2015fast,
sun2015cascaded,
tagliasacchi2015robust,
tang2015opening,
ge2016robust,
sinha2016deephand,
taylor2016efficient,
wan2016hand,
zhou2016model,
chen2017pose,
ge20173d,
guo2017towards,
oberweger2017deeppriorplus}. The same trend is also witnessed for the related problem of body pose estimation \cite{girshick2011efficient,sung2012unstructured,shotton2013real,sun2017compositional}. However, research on human body pose estimation has been continuously producing approaches that rely only on regular RGB  \cite{ferrari2008progressive,andriluka2008people,tompson2014joint,toshev2014deeppose,moreno20173d}.

The problem of 3D hand pose estimation and tracking based solely on color input has been studied for at least two decades~\cite{stenger2001model,sudderth2004visual,athitsos2003estimating,de2006regression,romero2009monocular,romero2010hands,oikonomidis2010markerless,de2011model,sridhar2014real,panteleris2017back,zimmermann2017learning} but has not seen an advancement that is comparable to that of human body pose estimation.
Earlier works on monocular RGB suffered from large  runtime and low accuracy. The work of Rehg and Kanade~\cite{rehg1994visual} required specialized hardware to achieve interactive performance at $10 Hz$ using stereo RGB input. Furthermore, the adopted tracking method did not support hand self-occlusions, limiting considerably the range of manageable hand motions. The work in \cite{de2011model} achieved accurate monocular RGB 3D hand tracking by modeling the observed scene using explicit parametric models for the hand, the lighting, and the background. A sophisticated optimization approach was developed to optimize the parameters, recovering the 3D hand pose. This explicit modeling and optimization however resulted in a computational cost that is prohibitive for real-time applications, even with the evolution of hardware since the introduction of the method in 2011.

\begin{figure*}[t]
\begin{center}
\includegraphics[width=0.95\textwidth]{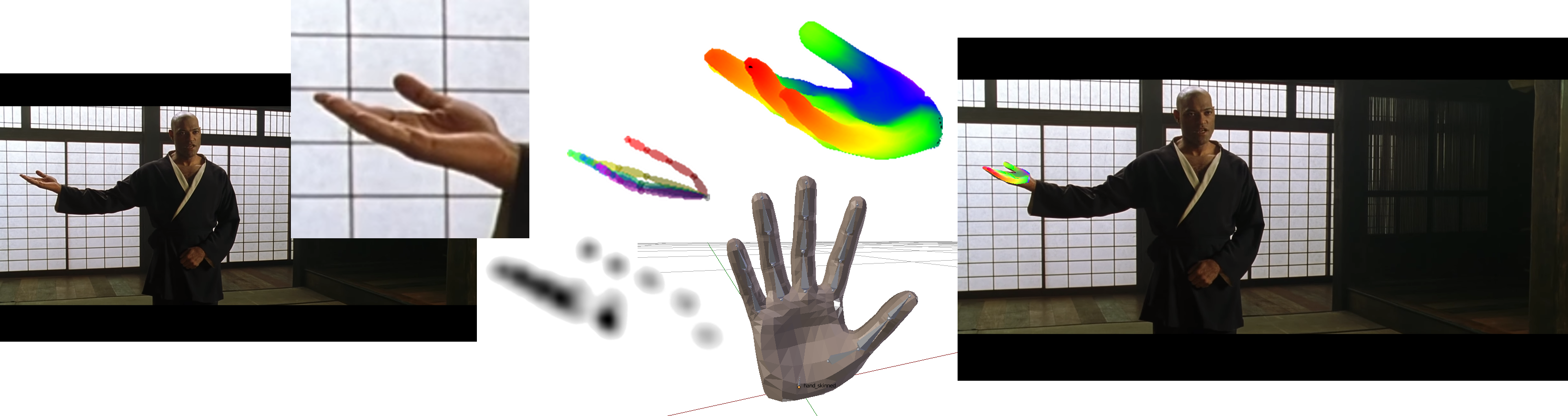}
\end{center}
\hspace*{2.8cm}(a) \hspace*{1.3cm} (b) \hspace*{1.0cm} (c) \hspace*{1.2cm} (d) \hspace*{1.0cm} (e) \hspace*{3.5cm} (f)
\caption{\label{fig:pipeline} Graphical illustration of the key steps of the proposed method. The method operates on a single RGB view (a). The hand is detected and a cropped image containing it (b) is provided to OpenPose that is responsible to estimate the 2D locations of hand joints (c). A hand model (d) is then transformed so that the distances of the projections of its joints to their observation counterparts are minimized (e). Figure (f) shows the final solution (e) superimposed to the input image (a).}
\end{figure*}

In our work, we capitalize upon the robust performance of state-of-the-art deep learning techniques and the power, versatility and  effectiveness of generative methods to achieve real-time 3D hand pose estimation that is robust enough to support real-world applications. 
A similar approach has been followed by~\cite{mehta2017vnect} but for the problem of 3D pose estimation of the human body. 
We adopt a three-step pipeline, namely hand detection, 2D key-point localization, and 3D pose estimation. For hand detection, we retrain a state-of-the-art detector~\cite{redmon2017yolo9000} to reliably detect hands in interactive frame rates. For the second step, we use the state-of-the-art 2D key-point localization approach from~\cite{simon2017hand}. Finally, for the 3D pose estimation step, we opt for a generative approach, allowing us to explicitly exploit known information such as personalized hand model metrics and camera calibration information. Therefore, our method estimates valid configurations of a hand whose dimensions do not vary over the frames of a sequence.

The closest to our approach is the one by Zimmermann and Brox~\cite{ zimmermann2017learning}
who also propose a method to estimate the 3D pose of a hand using as input a single color image. They also adopt a three-step pipeline as we do, but propose different solutions for each of these tasks. 
A striking difference between \cite{zimmermann2017learning} and our method, is that \cite{zimmermann2017learning} does not estimate the absolute 3D position of the hand, but rather a scale-normalized pose that is anchored on the palm center. Thus, its applicability in real world scenarios is limited.
Other advantages of our approach include its higher accuracy and its faster execution time. The first and last step are faster than the respective ones proposed by Zimmermann and Brox, while still allowing us to exploit the exceptional performance of the ``heavyweight'' architecture that         
OpenPose~\cite{simon2017hand} uses for our middle step.

Overall, to the best of our knowledge, the method proposed in this paper in the only one that tackles the problem of monocular, real-time 3D hand pose estimation robustly enough for real-world applications.

\section{3D hand pose from a single RGB frame}

The pipeline of the proposed method is illustrated in Figure~\ref{fig:pipeline}. Starting from a color image as input, we first detect all hands in it. For this task we employ a CNN-based object detector trained to identify hands in the RGB input. The architecture and training method of the network is described in Section~\ref{sec:detector}. For each hand in the input image, the detector produces a likelihood estimate and the coordinates of the hand's bounding box. We proceed to identify 2D joint positions on the detected hand(s) by passing each cropped hand image through a feed forward CNN. This produces heatmaps for the 2D joint locations of the hand which we convert to estimations of the 2D joint locations as described in Section~\ref{sec:joints_estimation}. The final step (Section~\ref{sec:ik}) lifts the 2D joints to 3D. We do so in a generative manner, employing a hand kinematics model and minimizing the distance of each projected model point from the corresponding 2D target point.

\subsection{Hand detector}
\label{sec:detector}
In the recent years, the state of the art in the task of object detection has progressed significantly with the use of convolutional neural networks. Many different architectures \cite{ ren2017faster,liu2016ssd, redmon2017yolo9000} have been proposed. These methods achieve impressive results in well-known benchmarks for hundreds or even thousands of object classes. They can also generalize well enough to perform ``in the wild''.

For the scope of this work, we are interested in just a limited number of classes and therefore, regarding specifically this aspect, most state-of-the-art methods are adequate. However, it is important to keep runtime low while also achieving a high quality detection rate with an as low as possible number of false positives. To this end we chose to follow the architecture of YOLO v2~\cite{redmon2017yolo9000}. YOLO is a state of the art object detector that has been demonstrated to outperform in detection accuracy more complex network architectures. At the same time, its runtime is low enough to enable its incorporation into realtime pipelines.

The YOLO V2 detector is an evolution of the older version presented in~\cite{redmon2016you}. The changes include the employment of the Darknet-19 architecture which resembles the VGG network architecture~\cite{simonyan2014very}. It is a fully convolutional architecture all the way to the last layer. It comprises a total of $19$ convolutional layers that along with the max pooling operations reduce the $288 \times 288$ input image to a score map of size $7 \times 7$.
Additional important changes include the use of batch normalization to speed-up and stabilize the training process, and a careful strategy to select anchor boxes.

A shortcoming of the selection of YOLO V2 in our application is the fact that the detection is performed in a local manner, not taking into account contextual information. This hinders the task of differentiating between left and right hands since the network has to decide the ``handedness'' of a region using the local image information, only. It is conceivable that this task would be improved if the detector took into account the image context.

Our hand detector was adapted from YOLO to have two classes: ``head'' and ``hand''. Having extra information about the location of body parts apart from the hands can be beneficial as an easy way to differentiate between left and right hand and to give clues regarding the tasks that the observed hands are engaged in.
We used the pre-trained weights for the convolutional layers from darknet19\footnote{\url{https://pjreddie.com/media/files/darknet19_448.conv.23}} as initialization for the retargeting process described below.

In order to train our detector we created a dataset, with a total of 13k RGB frames captured in VGA resolution. The dataset contains 12 subjects in different indoor environments. The subjects perform tasks such as typing on a computer, gesturing, and engaging in conversations. The frames were automatically annotated using OpenPose~\cite{cao2017realtime}. Before training the network, the dataset was split into a training set of about 12k and a validation set of about 1k frames. The network was trained for 20k iterations. The retrained network achieved a $92.8\%$ detection rate and $1.7\%$ false positive rate on the validation set.
Given an input image, the resulting detector can detect the two specified classes as well as their bounding boxes in constant time.

\subsection{2D joints estimation}
\label{sec:joints_estimation}
In the last few years a lot of research has been devoted to methods that employ CNN based architectures to perform human body 2D joint estimation in RGB images. 
These methods capitalized on the emergence of large datasets with annotated human body parts and poses such as the MS COCO~\cite{mscoco} and the MPII~\cite{mpii}. 
As a result, the state of the art on this task has moved forward significantly. Nevertheless, mainly due to the lack of annotated datasets, the body-keypoint architectures could not be applied for hand joint estimation. Only recently, the work of Simon et al~\cite{simon2017hand} used multiview bootstrapping in order to iteratively create a large real world dataset of hands annotated with 2D joint positions. The multiview dataset enables the annotation of all joints on each frame, even if they are occluded. This, in turn, enables to train a network with an architecture similar to Wei's et al.~\cite{wei2016} that learns to estimate visible as well as occluded joints.
OpenPose achieves state of the art results in difficult datasets and is shown to perform in the wild. This ability to generalize and produce good 2D keypoint detections from arbitrary hand images is key to the goals of this work. 

We incorporated the work of~\cite{simon2017hand} into our pipeline in order to detect the 2D positions of hand joints. More specifically, we crop the image according to the detected bounding box (see Section~\ref{sec:detector}) and feed it to the 2D keypoint detector. Since the keypoint detector is only trained on left hands, we handle right hands by first mirroring the image along the Y axis. The output of the detector is $21$ heatmaps that correspond to estimates for the 20 hand keypoints (four per finger) and one wrist point.

\subsection{From 2D joints to 3D pose}
\label{sec:ik}
The task of computing the 3D hand pose using 2D joint estimations from a single view is formulated as an inverse kinematics (IK) problem. 

The information about the scale of the observed object is lost during the projection transformation. To account for that, we use a 3D hand model that is close in scale to the observed hand. Apart from the scale, the 3D hand model constrains the solution space to only plausible hand articulations by explicitly encoding joint limits.

\subsubsection{Hand model}
The hand model has $26$ degrees of freedom represented by $27$ parameters, similar to ~\cite{oikonomidis2011efficient} and others. The global translation and rotation of the hand requires $6$ degrees of freedom (DoFs), encoded by $7$ values since we adopt the representation of quaternions for 3D rotations. The joint at the base of each finger is modeled using two DoFs and the rest of the finger joints require one DoF each. The finger joints are bound by the joint limits that apply to a real hand. 
We identify keypoints on the model skeleton that correspond to the locations of the joints that the 2D joint detector estimates.

\subsubsection{Single camera view}
\label{sec:mono}
Given a hand pose $P$ defined by the $27$ parameters and the forward kinematics function for the hand model $F$, we compute the 3D positions of the joint keypoints using $K^P = F(P)$ in the world coordinate frame. 

By applying the camera view matrix $C_{v}$ and the camera projection matrix $C_{p}$, we first transform the points to the coordinate frame of the camera and then project them on the camera plane:
\begin{equation}
\label{eq:proj}
M_i^P = C_{p} \cdot C_{v} \cdot K_i^P,
\end{equation}
where $M_i^P$ is the projection $(x_i, y_i)$ of the joint $i$ on the image plane.
In the case a single view/camera is used, we can always choose to keep the model in the camera frame. In this case $C_{v}$ is the identity matrix. 

Let $ J_i = (u_i, v_i, p_i)$,  $i \in [1,21]$, represent the $21$ detected 2D hand joints (see Section~\ref{sec:joints_estimation}). 
$(u_i, v_i)$ are the 2D coordinates of the joint on the input image and $p_i$ is the network's confidence for the joint $i$, ($p_i \in [0,1]$). 
In order to avoid using false detections in the IK step we do not consider joints $J_i$ with confidence $p_i$ below an experimentally identified value  $p^{th} = 0.1$.

\begin{figure}[t]
\begin{center}
\includegraphics[width=0.35\textwidth]{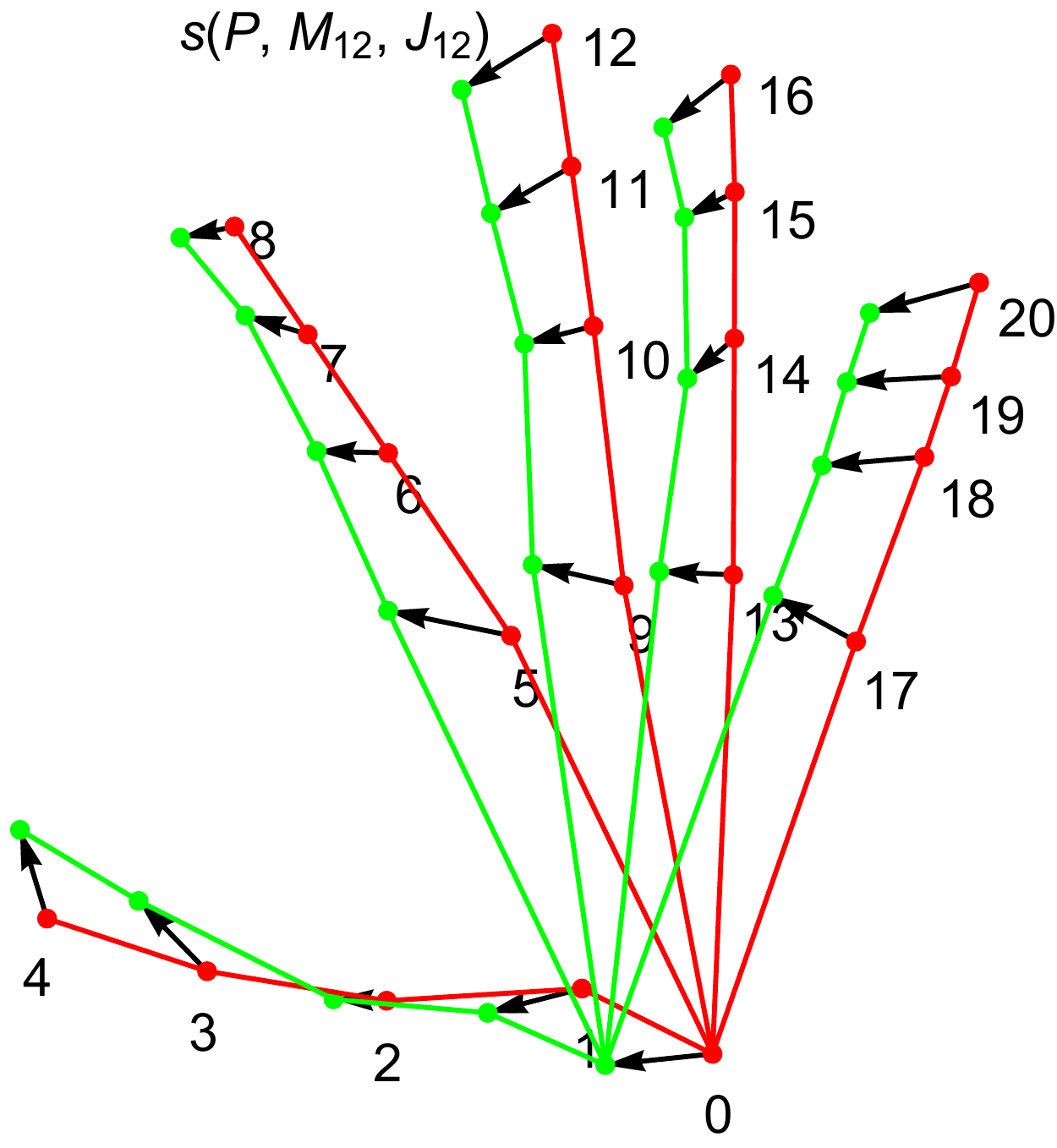}
\caption{Illustration of the residuals between the observed hand joints (green) and the model keypoints (red) described in Eq.(\ref{eq:totalscore}).}
\label{fig:residuals}
\end{center}
\end{figure}

For a given pose $P$, we quantify the discrepancy $s(P, M_i, J_i)$ between the observed joint $J_i$ and the computed one $M_i$ as:
\begin{equation}
\label{eq:score}
    s(P, M_i, J_i) = (p_i^3 \cdot (x_i - u_i))^2  +  (p_i^3 \cdot (y_i - v_i))^2 .
\end{equation}
Similarly, the total discrepancy $S(P,M,J)$ between the observed and model joints can be computed as:
\begin{equation}
\label{eq:totalscore}
    S(P,M,J) = \sum \limits_{i=0}^{21} s(P, M_i, J_i).
\end{equation}
The 3D hand pose $P^*$  that is most compatible with the available observations (observed 2D joints) can be estimated by minimizing the objective function of Eq.(\ref{eq:totalscore}):
\begin{equation}
\label{eq:minscore}
    P^* = \arg \min_P \{ S(P,M,J) \}.
\end{equation}
This is achieved by using the Levenberg-Marquardt optimizer that minimizes this objective function after the automatic differentiation of the residuals.

Figure~\ref{fig:residuals} illustrates graphically the employed objective function. The hypothesized keypoints are shown in red, and the target ones in green. The sum of squares of the arrow lengths forms our objective function to be minimized, subject to the kinematic constraints of the hand.

\subsubsection{Stereo or multicamera input}
\label{sec:stereo}
The presented pipeline can be extended to support multiple input cameras (i.e, stereo) in a straightforward manner. More specifically, after the 3D keypoint generation, Eq.(\ref{eq:proj}) is applied separately to each camera, using the corresponding $C_{v}$ and $C_{p}$. This yields a set of 2D keypoints per camera, corresponding to the ones detected by OpenPose on the respective camera image. We proceed to formulate our non-linear least squares problem by defining and minimizing the sum of 2D residuals across all views:
\begin{equation}
{\cal S}_{all}(P) = \sum \limits_{c \in C} S(P_c , M_c , J),
\end{equation}
where $C$ is assumed to be the set of available cameras. Experimental results (see Section~\ref{sec:error}) show that the availability of more views improves hand pose estimation accuracy considerably as additional constraints are provided to the IK problem.

\begin{figure*}[t]
\begin{center}
\includegraphics[width=0.32\textwidth]{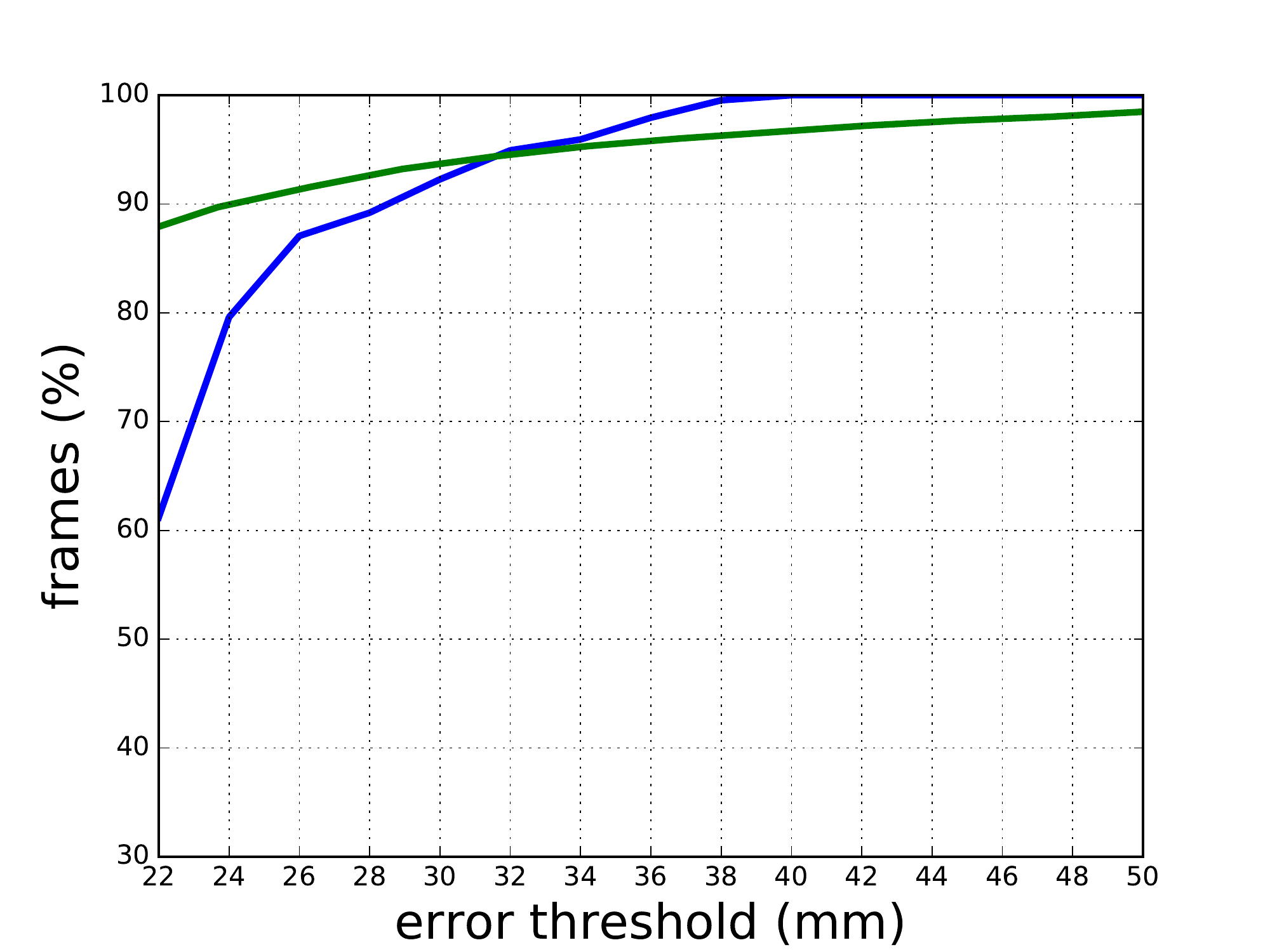}
\includegraphics[width=0.32\textwidth]{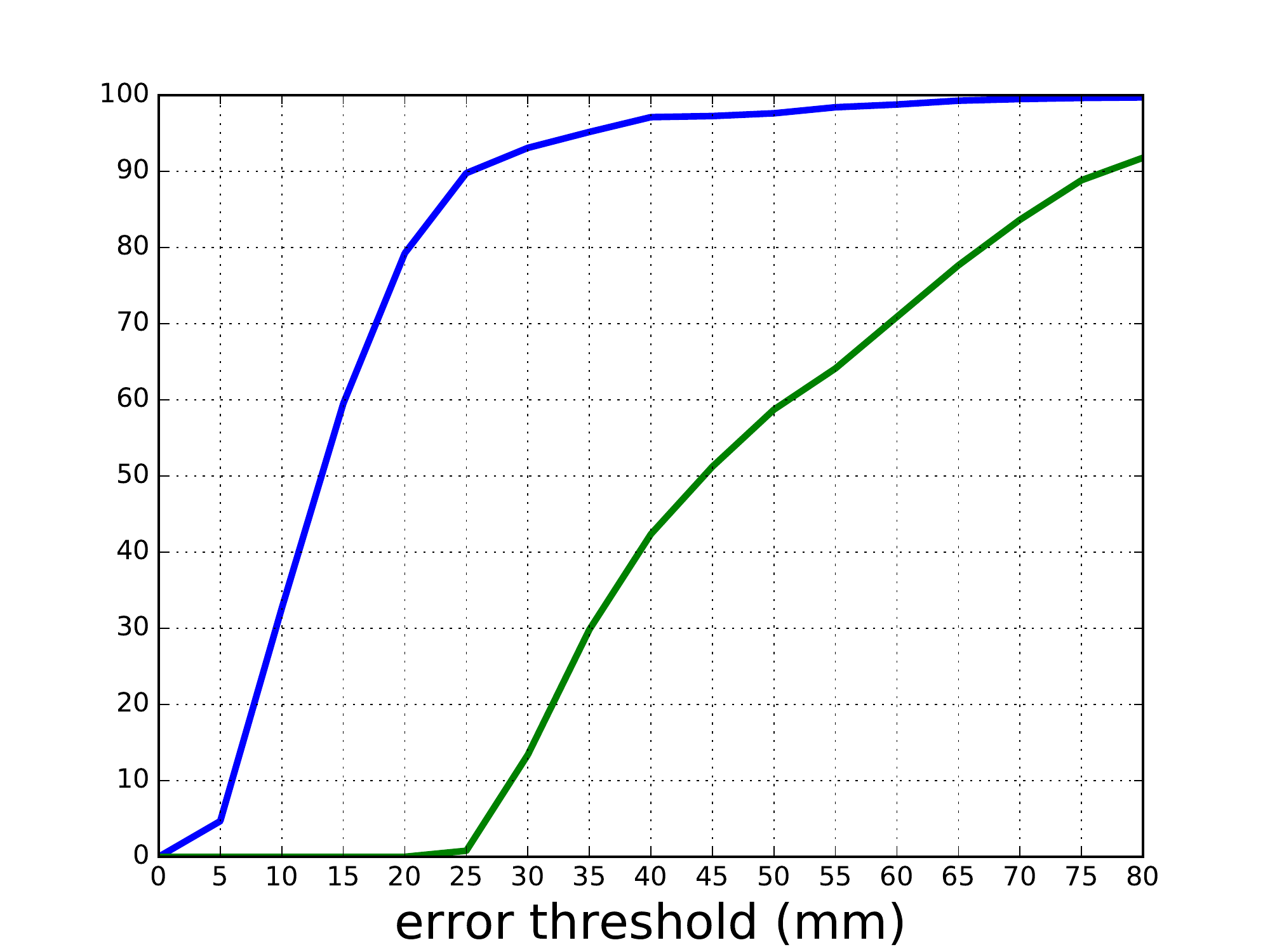}
\includegraphics[width=0.32\textwidth]{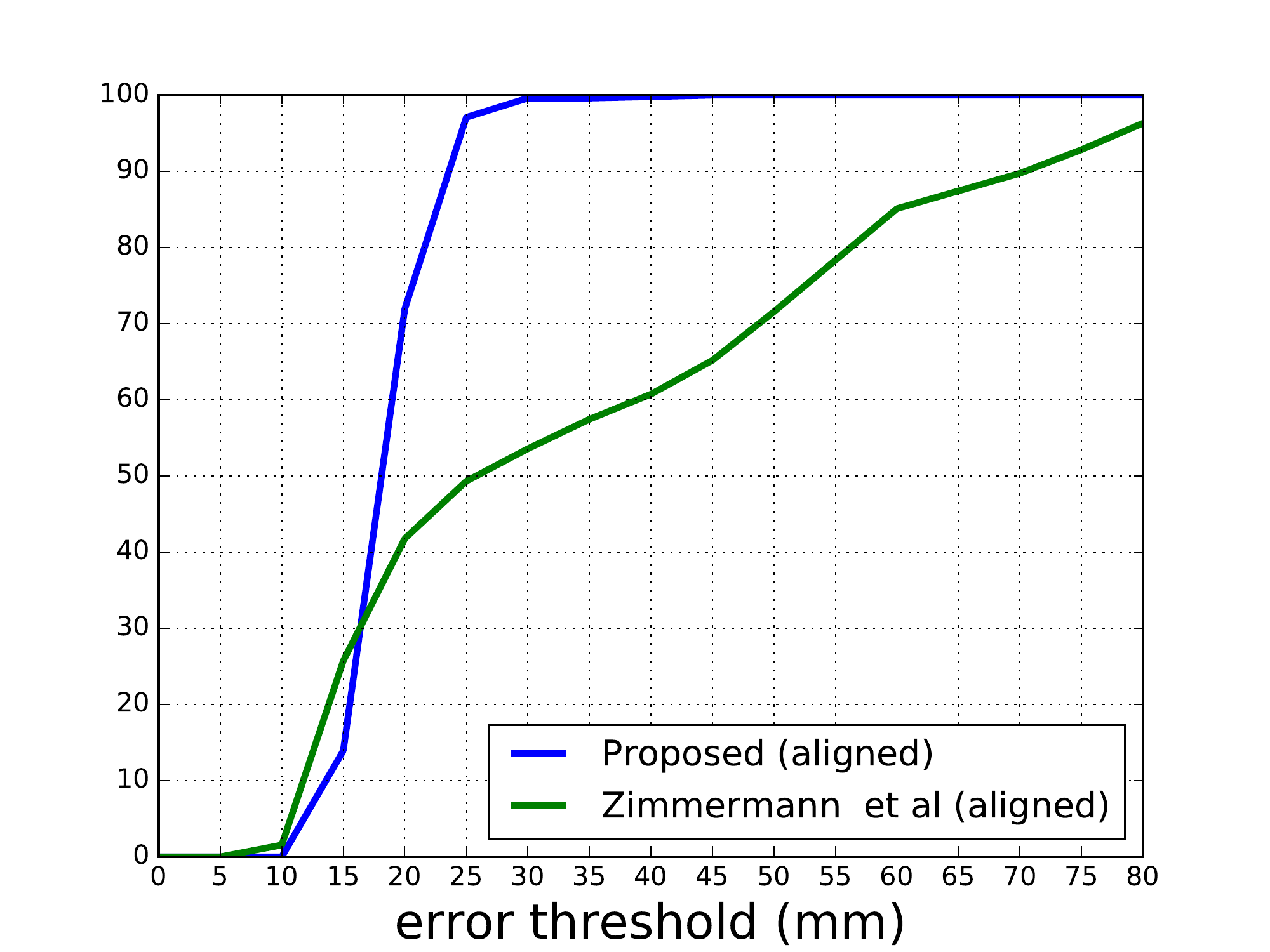}
\caption{Percentage of frames ($y$ axis) for which the average 3D joints estimation error is below a certain threshold ($x$ axis). We plot and compare the performance of the proposed method (blue curves) compared to the method of Zimmermann and Brox~\cite{zimmermann2017learning} (green curves) in the {\bf SHD}~\cite{zhang20163d} (left),  {\bf B2RGB}~\cite{panteleris2017back} (middle) and {\bf HIC}~\cite{tzionas2015capturing} (right) datasets.}
\label{fig:zim}
\end{center}
\end{figure*}

\subsection{Implementation details}
The proposed method assumes that the intrinsic parameters of the camera (focal length, camera center, distortion) are known. 
In our reference implementation we used three different sets of left and right hand models. The first set was created with the makehuman\footnote{\url{http://www.makehuman.org/}} 3D character creation tool. The second was adapted from the libhand\footnote{\url{https://github.com/libhand/libhand}} project. The libhand model is an anatomically accurate skinned hand model with a skeleton that supports realistic articulations. 
The third hand model is a scaled version of the makehuman model used to compare our method with Zimmermann's et al.~\cite{zimmermann2017learning} on the Zhang et al.~\cite{zhang20163d} dataset (see Section~\ref{sec:quantitative} and Figure~\ref{fig:zim} (left)).

The proposed method operates on a single frame, without requiring any form of initialization. However, when employed on a video where the assumption of temporal continuity holds between several frames, it is useful to start the IK optimization step from the last known solution, i.e., use the method in a tracking mode. However, this does not prevent the method to operate correctly even in the case of abrupt hand motions, provided that OpenPose gives reliable estimates of the 2D locations of joints.

In certain cases, when joints become invisible, OpenPose reports 2D locations with very low confidence. In such situations, we consider the last known position of that joint and transform it according to the rotation $R$ and translation $T$ of the hand root pose.

In our implementation, optimization has been performed by employing the Ceres Solver~\cite{ceres}. 

For the single handpose estimation scenario the proposed method is able to perform in real-time. Tested on a workstation with an Intel i7 CPU and an NVIDIA GTX 1070 GPU, our reference implementation achieves 18fps. In that setup, the majority of the processing time, per frame, is consumed by OpenPose with about ~30ms, while the hand detection step requires about 16ms.

\section{Experiments}
\label{seq:experiments}

We evaluated experimentally our method both quantitatively and qualitatively and compared it to~\cite{zimmermann2017learning}.
\subsection{Selected datasets}
\label{sec:quantitative}
Despite the importance of the 3D hand pose estimation problem, there are surprisingly few datasets available that can be used for benchmarking and comparing different methods. This scarcity is mainly due to the difficulty of acquiring accurate ground truth without the use of intrusive methods like markers or sensor gloves. Since the advent of the RGB-D sensors, most of the methods for 3D hand pose estimation relied on depth data. There is a number of datasets available which provide manual or semi-automatic annotations on depth data. In recent years, the NYU hand pose dataset~\cite{tompson2014real} has become a standard dataset for depth-based methods. The ground truth for this dataset is computed using a generative method~\cite{oikonomidis2011efficient} instead of manual labeling. Unfortunately, the creators of this dataset provide RGB images that are warped onto the depth map. As it is also pointed out by Zimmermann and Brox~\cite{zimmermann2017learning}, this makes the dataset unusable by RGB-only methods. In the same work, Zimmermann proposed a new synthetic dataset for handpose estimation. Unfortunately they rendered the synthetic poses in low resolution ($320 \times 320$) which makes it unusable by our method. The dexter1~\cite{sridhar2013interactive} dataset would fit our requirements but as it is pointed out by Simon et al~\cite{simon2017hand} it suffers from bad calibration and synchronization issues. Finally the datasets by Gomes et al~\cite{gomez2017large} and Oberweger et al~\cite{oberweger2016efficiently} do not have 3D ground truth available yet.

Thus, in our work we used three publicly available datasets, the Stereo handpose dataset~\cite{zhang20163d} ({\bf SHD}), a synthetic dataset presented in~\cite{panteleris2017back} ({\bf B2RGB}) and the hands in action RGB-D dataset \cite{tzionas2015capturing} ({\bf HIC}).

\paragraph{Stereo handpose dataset~\cite{zhang20163d} (SHD):} {\bf SHD} is used for training and quantitative evaluation in ~\cite{zimmermann2017learning}. We choose to include this dataset in our quantitative evaluation in order to provide a direct comparison with~\cite{zimmermann2017learning}, even though the accuracy of the provided ground truth is limited. 

\paragraph{Synthetic dataset \cite{panteleris2017back} (B2RGB):} This dataset contains sequences of a realistic hand model from {\sl libhand}. 
The sequences include articulations of a single hand, hand-object interaction and two strongly interacting hands. The same sequences are available rendered from a virtual stereo pair as well as a virtual RGB-D sensor. Being synthetic, this dataset is annotated with perfectly accurate ground-truth.

\paragraph{Hands in action RGB-D dataset \cite{tzionas2015capturing} (HIC):} {\bf HIC}
consists of multiple RGB-D sequences of one or two hands performing articulations. The authors of the dataset provide manual 2D annotations of the sequences at intervals of $5$ frames. Additionally, the dataset is accompanied with the tracks of the hands, as those were estimated by~\cite{tzionas2015capturing}. 

\subsection{SOTA comparison}
Figure~\ref{fig:zim} shows the percentage of frames ($y$ axis) for which the average 3D joints estimation error is below a certain threshold ($x$ axis). We plot and compare the performance of the proposed method (blue curves) compared to the method of Zimmermann and Brox~\cite{zimmermann2017learning} (green curves) in the {\bf SHD}~\cite{zhang20163d} (left),  {\bf B2RGB}~\cite{panteleris2017back} (middle) and {\bf HIC}~\cite{tzionas2015capturing} (right) datasets.
It should be stressed that the method of Zimmermann does not estimate the 3D location of the hand or the scale of the estimated hand. Therefore, in Figure~\ref{fig:zim} we report output that is processed similarly to the final 3D output in~\cite{zimmermann2017learning} (labeled ``proposed aligned''). For a fair comparison, the output of our method is also post-processed in exactly the same way. As it can be observed, our method outperforms largely the method of Zimmermann in the {\bf B2RGB} and the {\bf HIC} datasets. In the {\bf SHD} dataset, the method of Zimmermann performs better for error thresholds less than $30$mm. This is attributed to the fact that their CNN-based approach is trained on this dataset. 
Additionally, the low performance of our method below $30 mm$, is attributed to the difference in hand dimensions between our hand model and the human hand in {\bf SHD}.

\begin{figure}[t]
\begin{center}
\includegraphics[width=0.95\columnwidth]{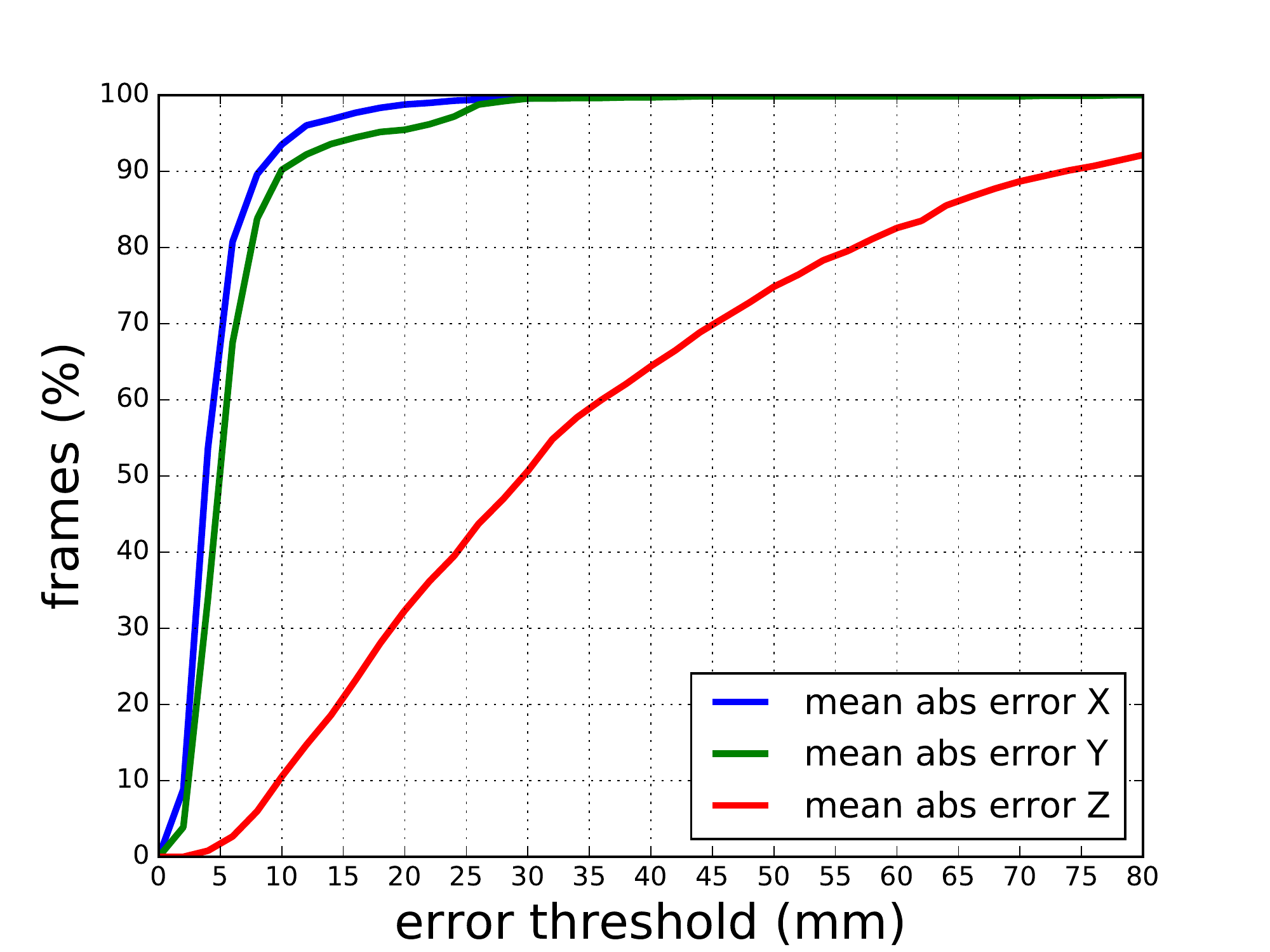}
\caption{Percentage of frames ($y$ axis) for which the average 3D joints estimation error is below a certain threshold ($x$ axis). The plot shows the per axis, mean absolute error of the proposed method for the {\bf B2RGB} dataset. Due to the uncertainty of the monocular view the error is mainly found along the Z axis.}
\label{fig:axiserror}
\end{center}
\end{figure}

\begin{figure}[t]
\begin{center}
\includegraphics[width=0.92\columnwidth]{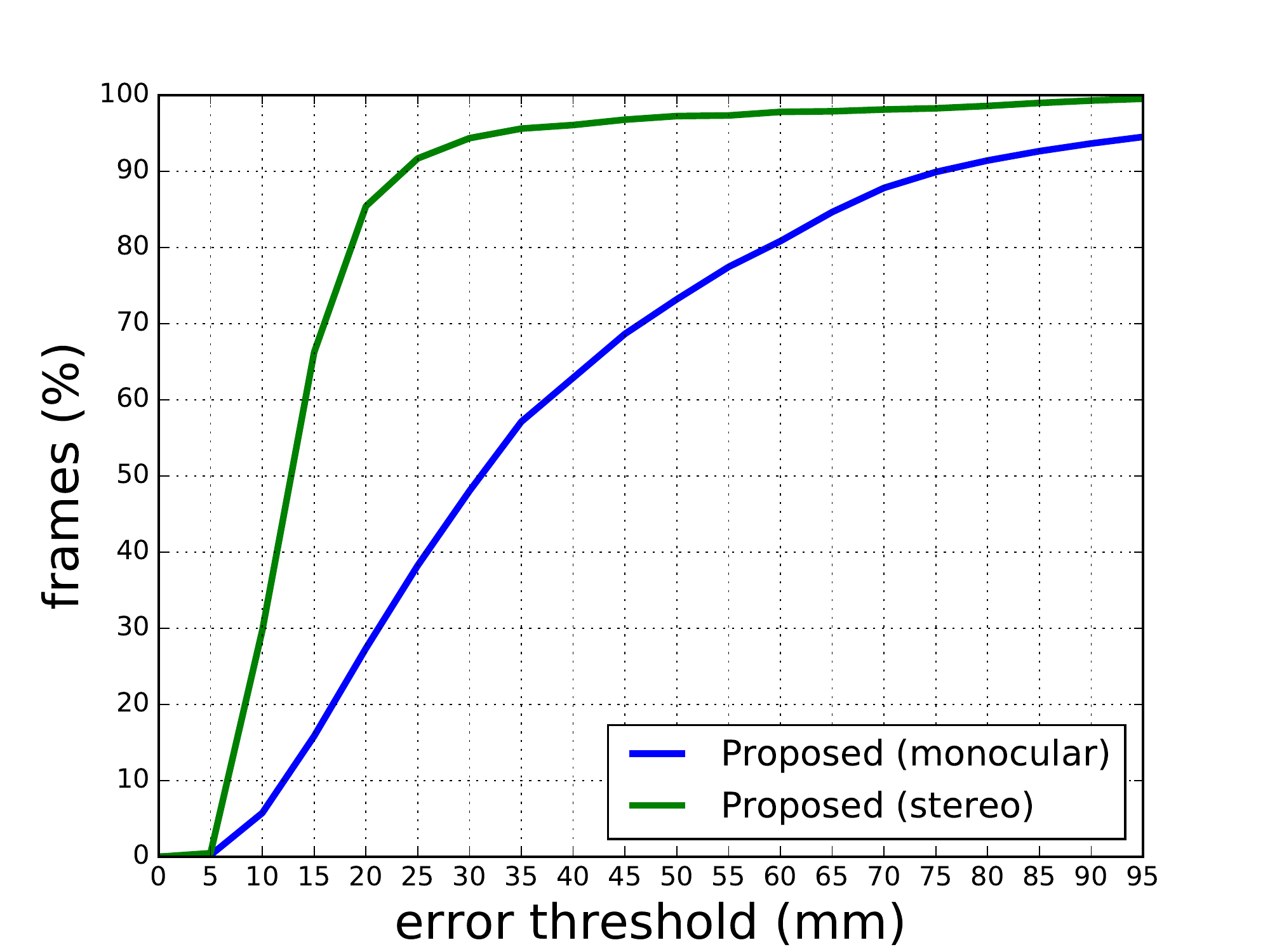}
\caption{Percentage of frames ($y$ axis) for which the average 3D joints estimation error is below a certain threshold ($x$ axis). We plot the performance of the proposed method applied to monocular input (Section~\ref{sec:mono}) in comparison to the case where a stereo input is used (Section~\ref{sec:stereo}) on the {\bf B2RGB} dataset~\cite{panteleris2017back}. It can be verified that the addition of a second camera increases considerably the accuracy in 3D hand pose estimation.}
\label{fig:monovsstereo}
\end{center}
\end{figure}

\subsection{Error analysis}
\label{sec:error}
The per-axis error of the proposed method is shown in Figure~\ref{fig:axiserror}. It is evident that the main source of uncertainty is the depth estimation ($Z$ axis), while the accuracy for the estimation on the $X$ and $Y$ is below $10$mm for $90\%$ of the frames. This observation is further supported when we compare with the accuracy achieved for the same dataset using stereo input. 
Figure~\ref{fig:monovsstereo} illustrates the performance of the proposed method in case a single RGB view is used  
(Section~\ref{sec:mono}) in comparison to the case where stereo input is used (Section \ref{sec:stereo}) on the {\bf B2RGB} dataset. It can be verified that the addition of a second camera increases considerably the accuracy in 3D hand pose estimation.

\begin{figure*}[t]
\begin{center}
\includegraphics[width=0.49\columnwidth]{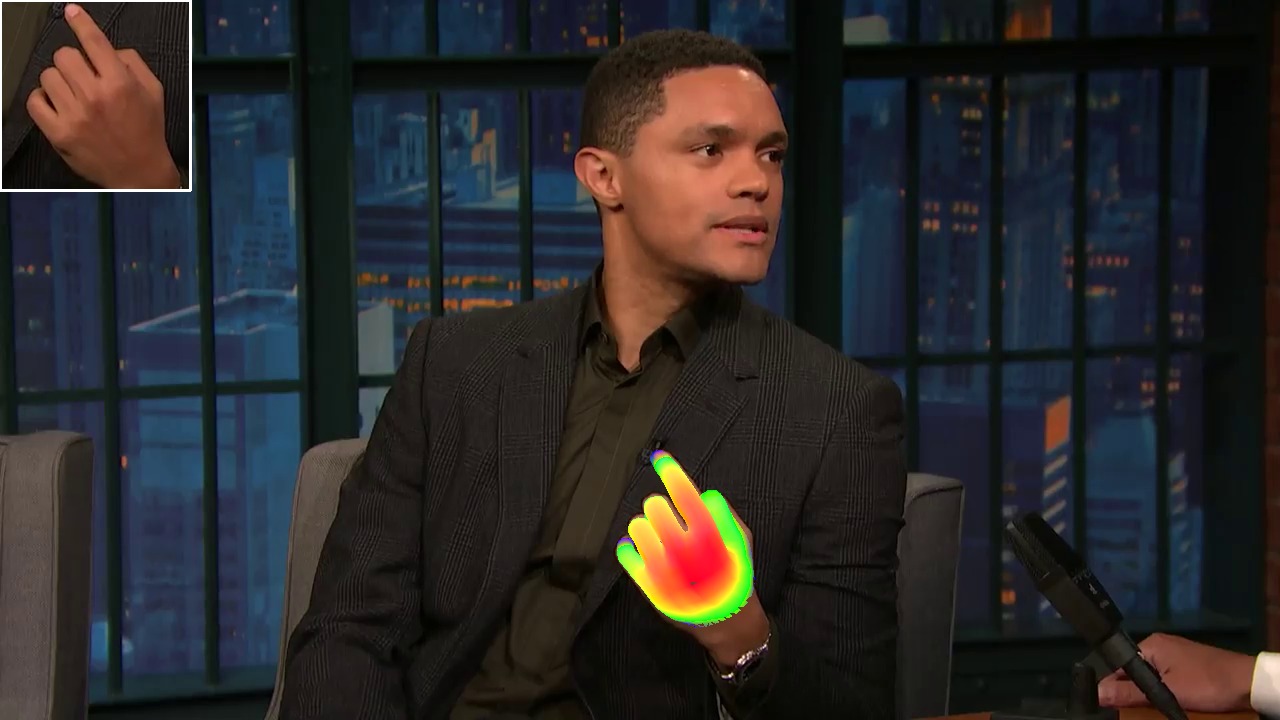}
\includegraphics[width=0.49\columnwidth]{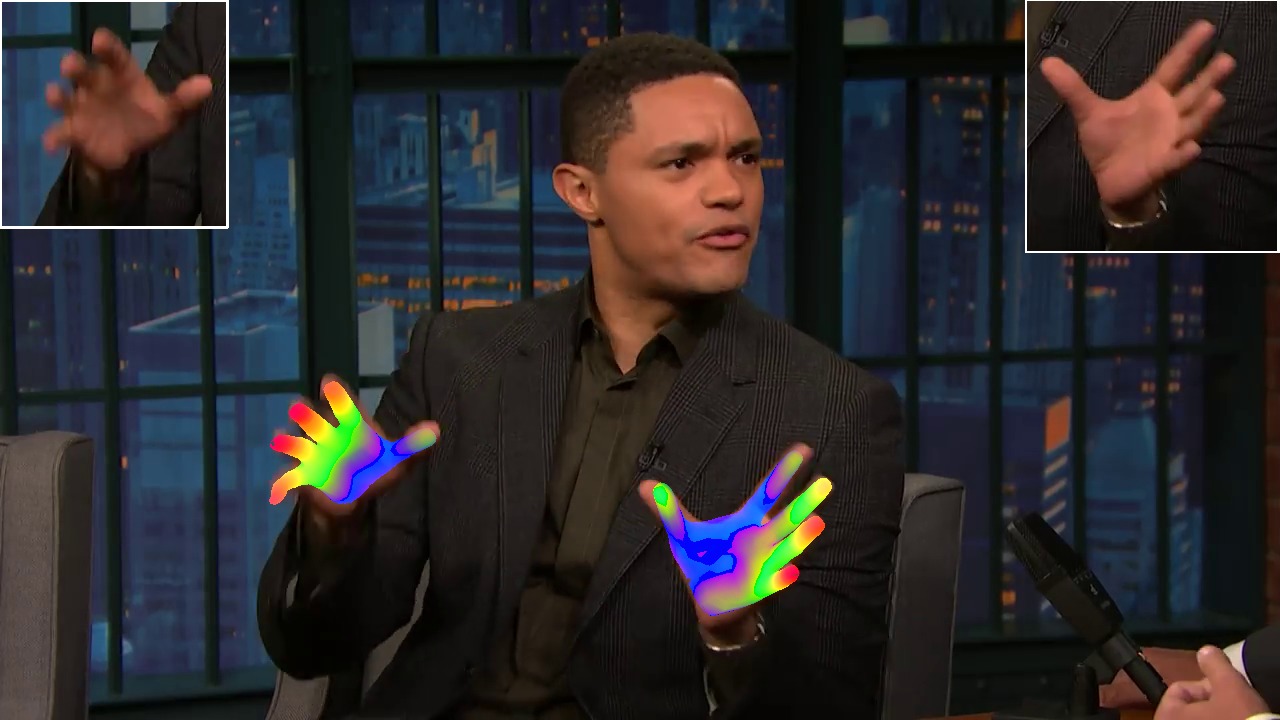}
\includegraphics[width=0.49\columnwidth]{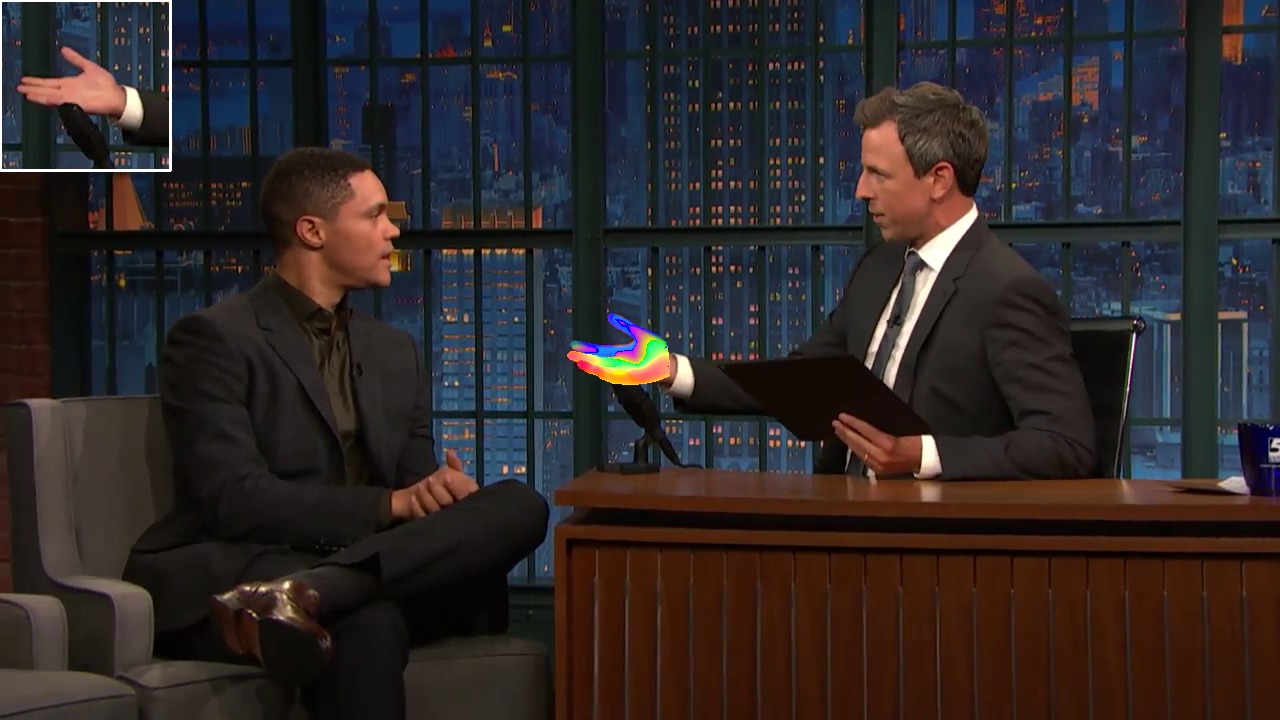}
\includegraphics[width=0.49\columnwidth]{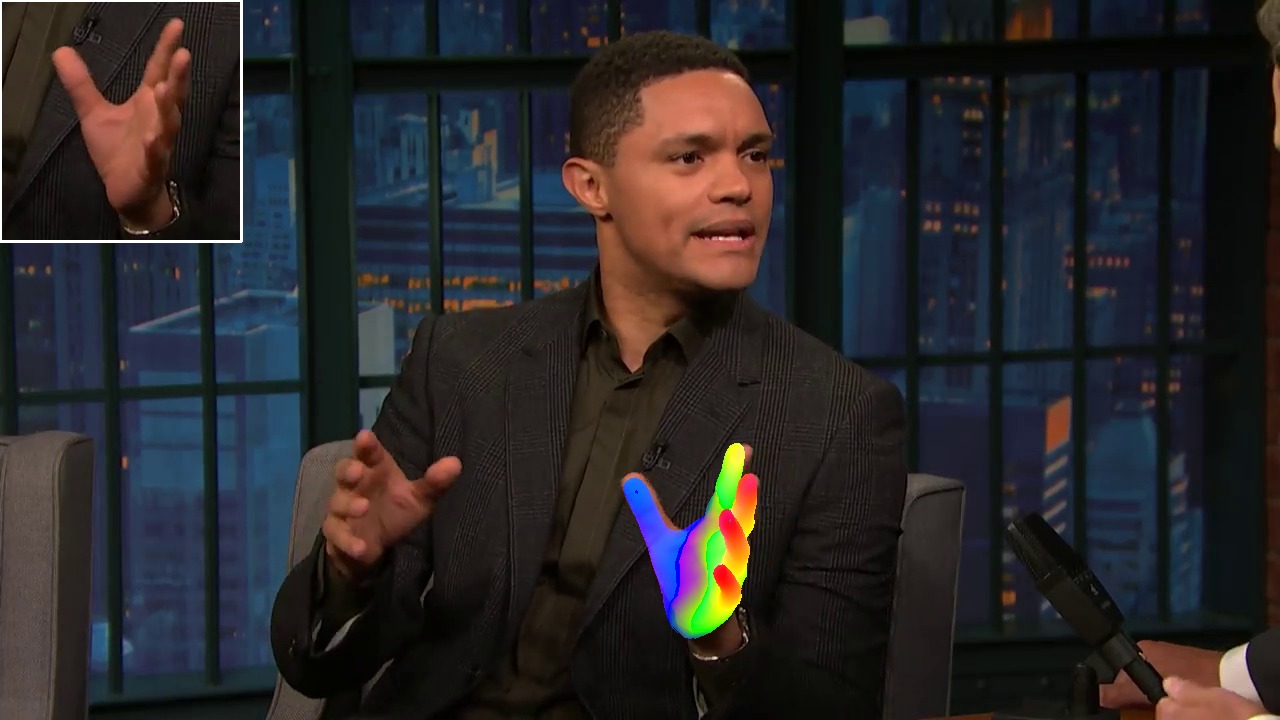}\\
\includegraphics[width=0.49\columnwidth]{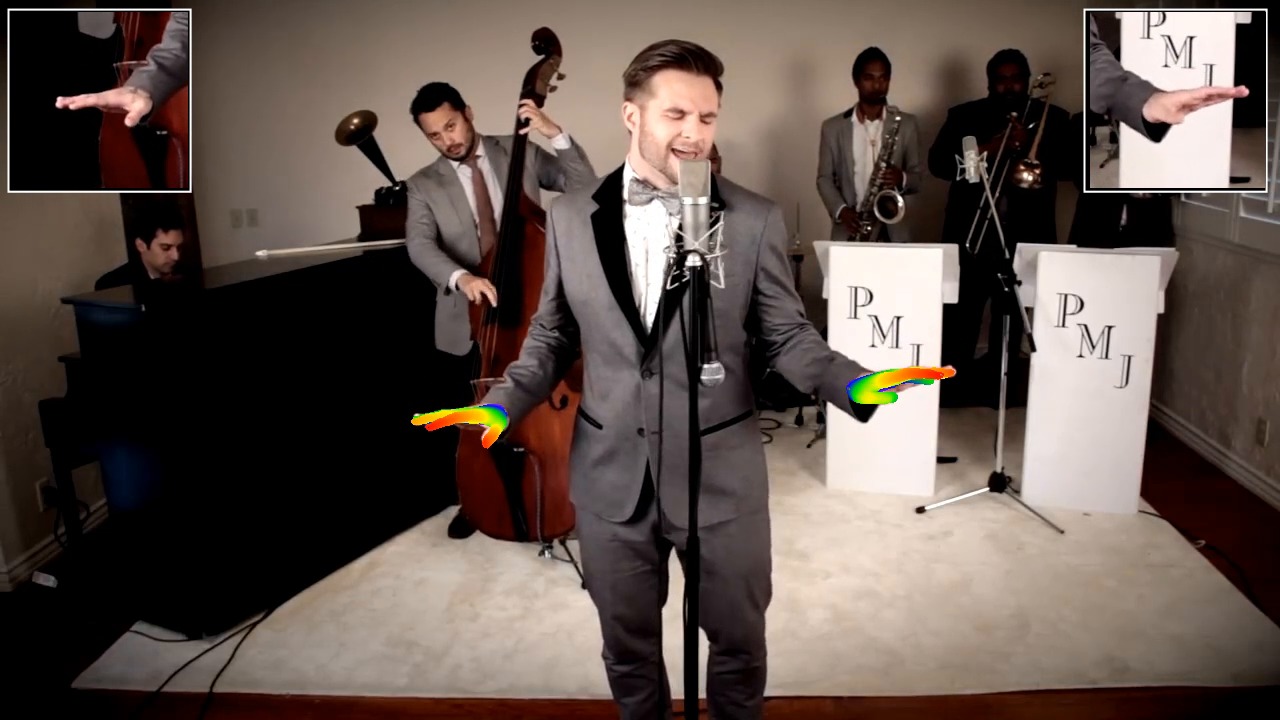}
\includegraphics[width=0.49\columnwidth]{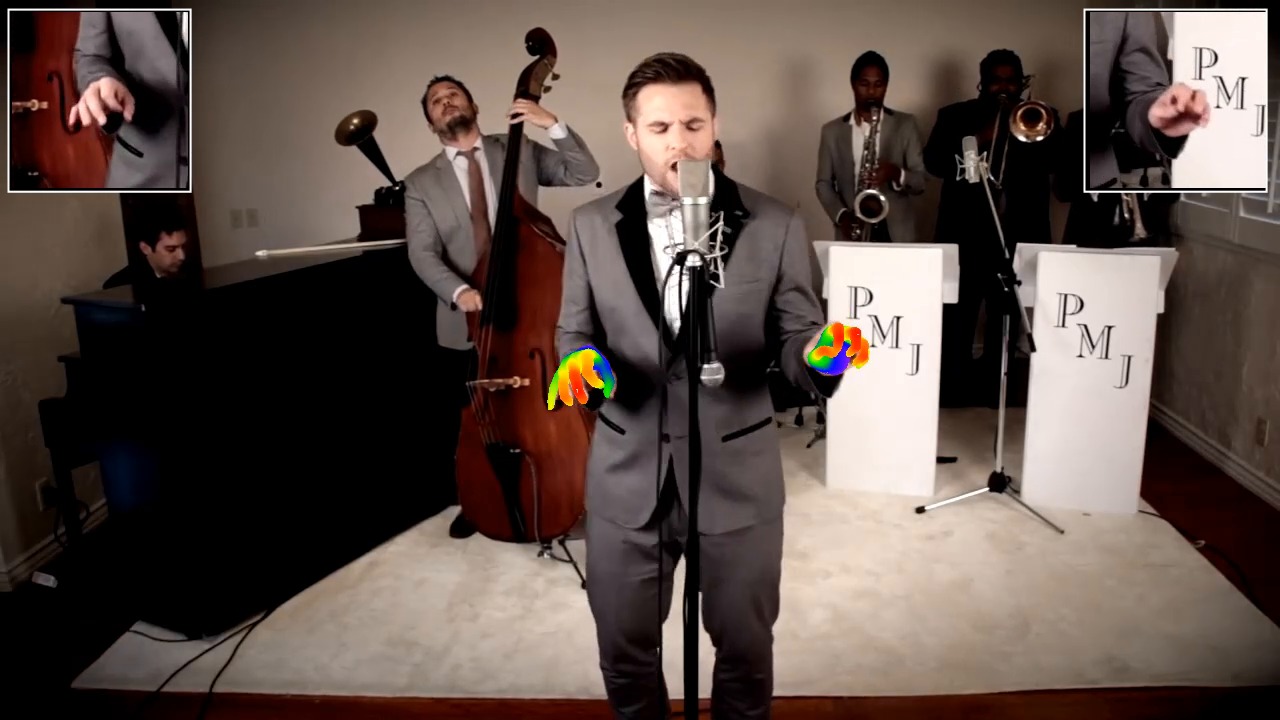}
\includegraphics[width=0.49\columnwidth]{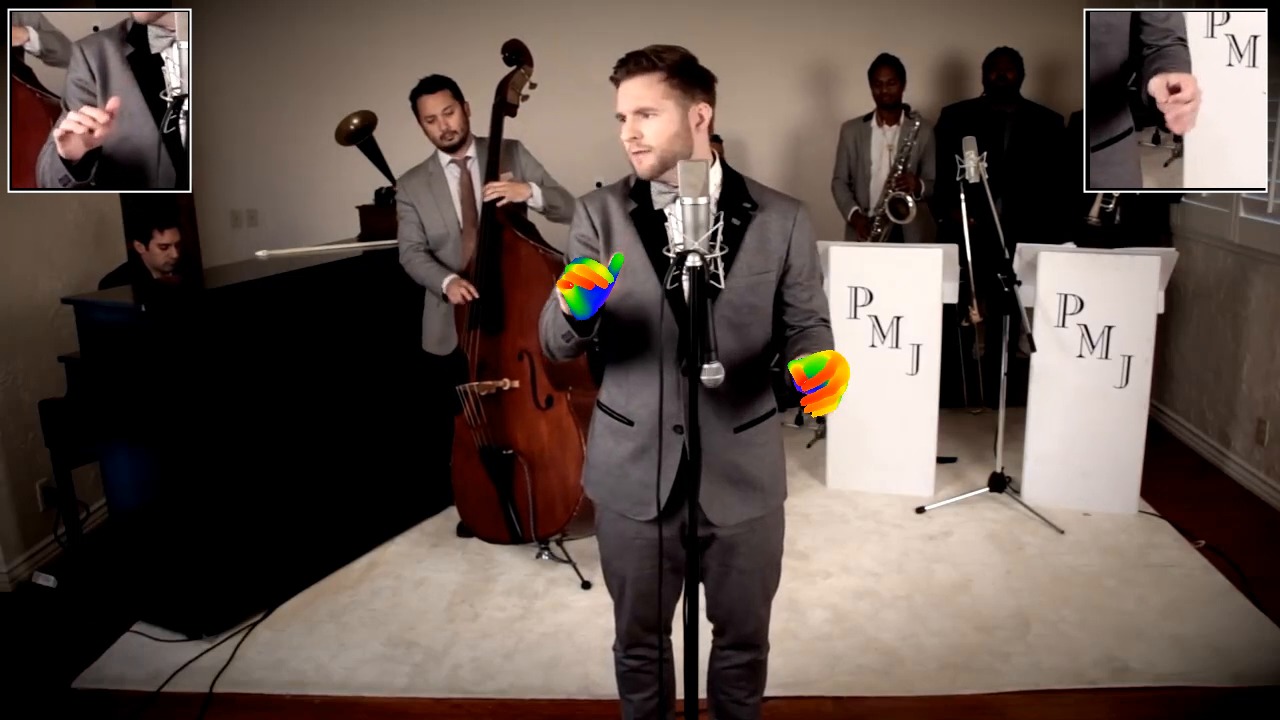}
\includegraphics[width=0.49\columnwidth]{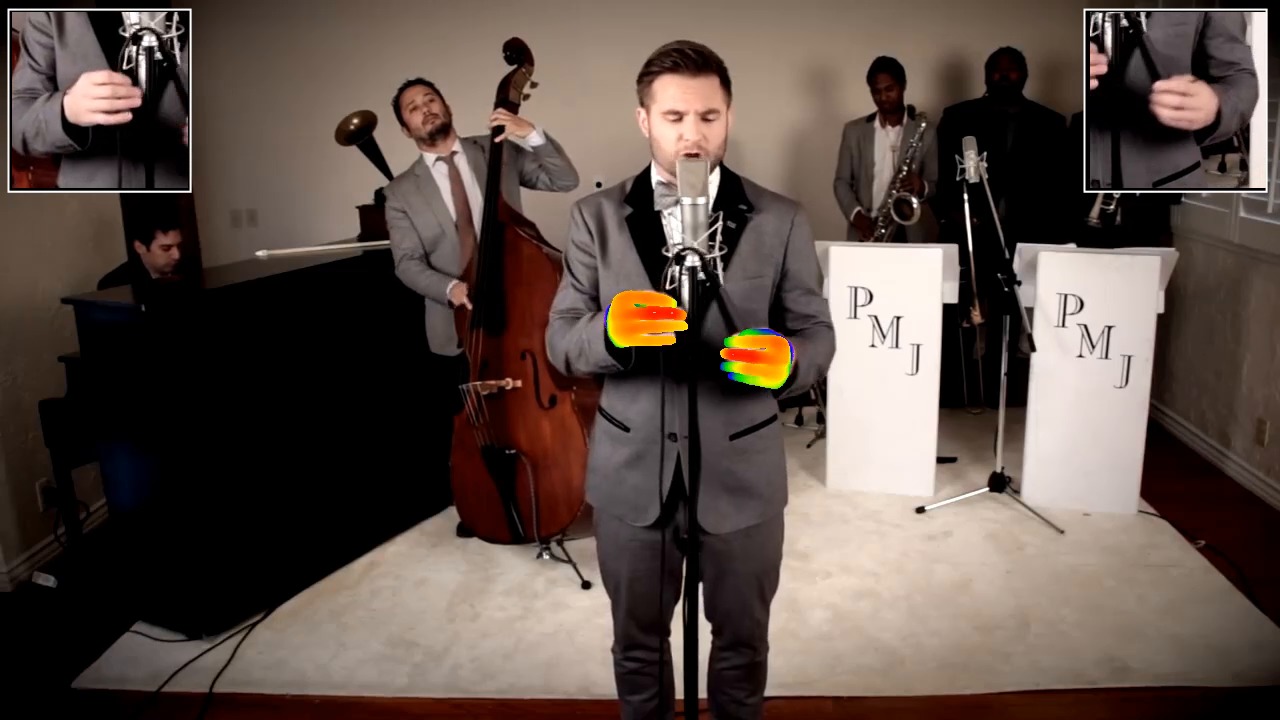}\\
\includegraphics[width=0.49\columnwidth]{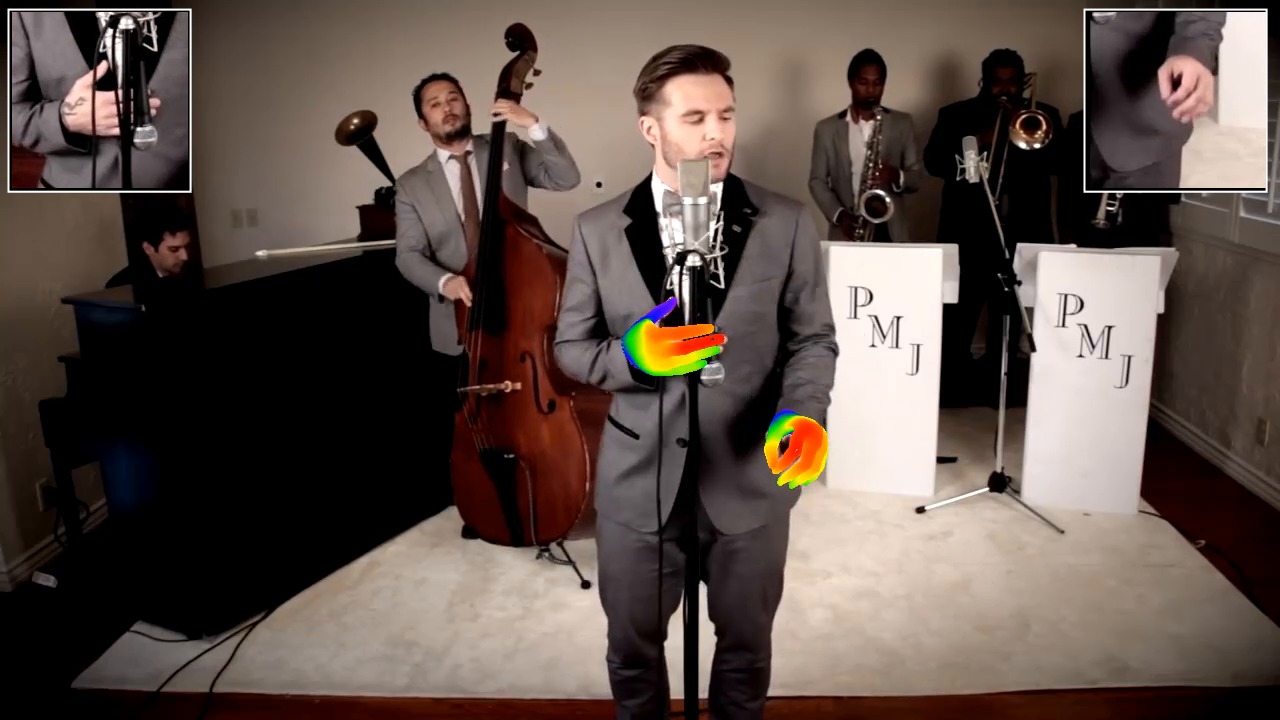}
\includegraphics[width=0.49\columnwidth]{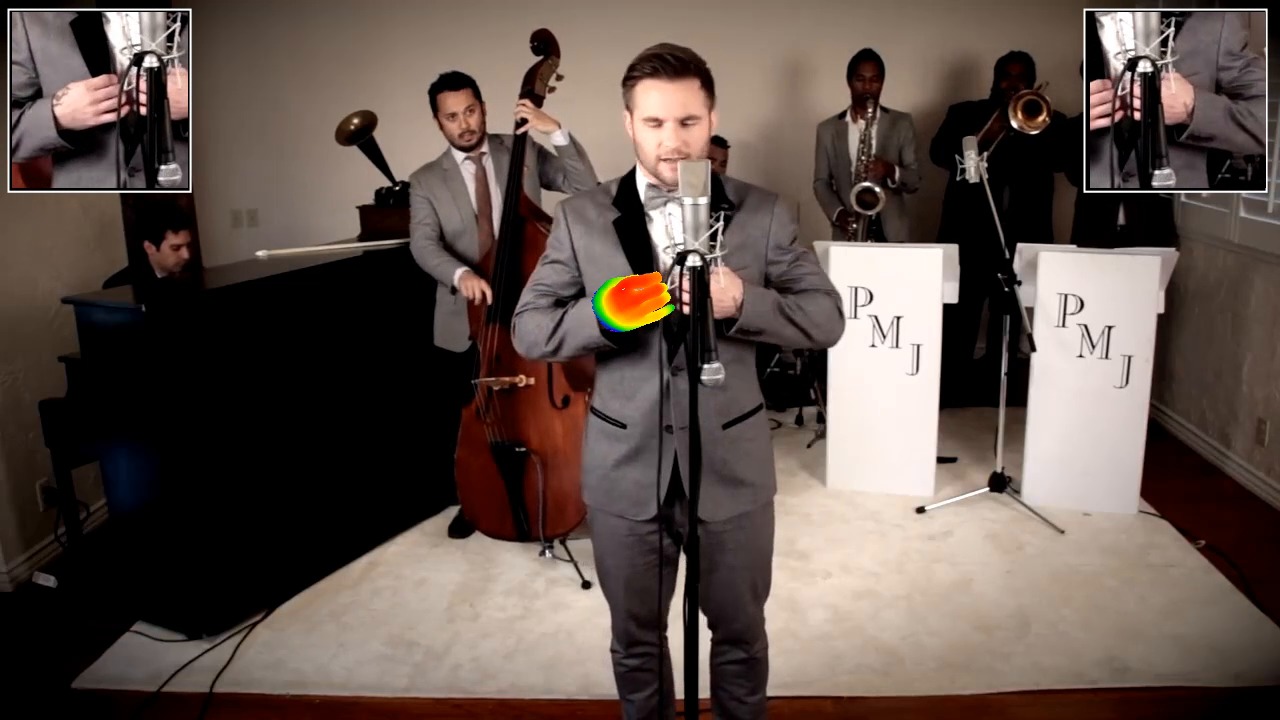}
\includegraphics[width=0.49\columnwidth]{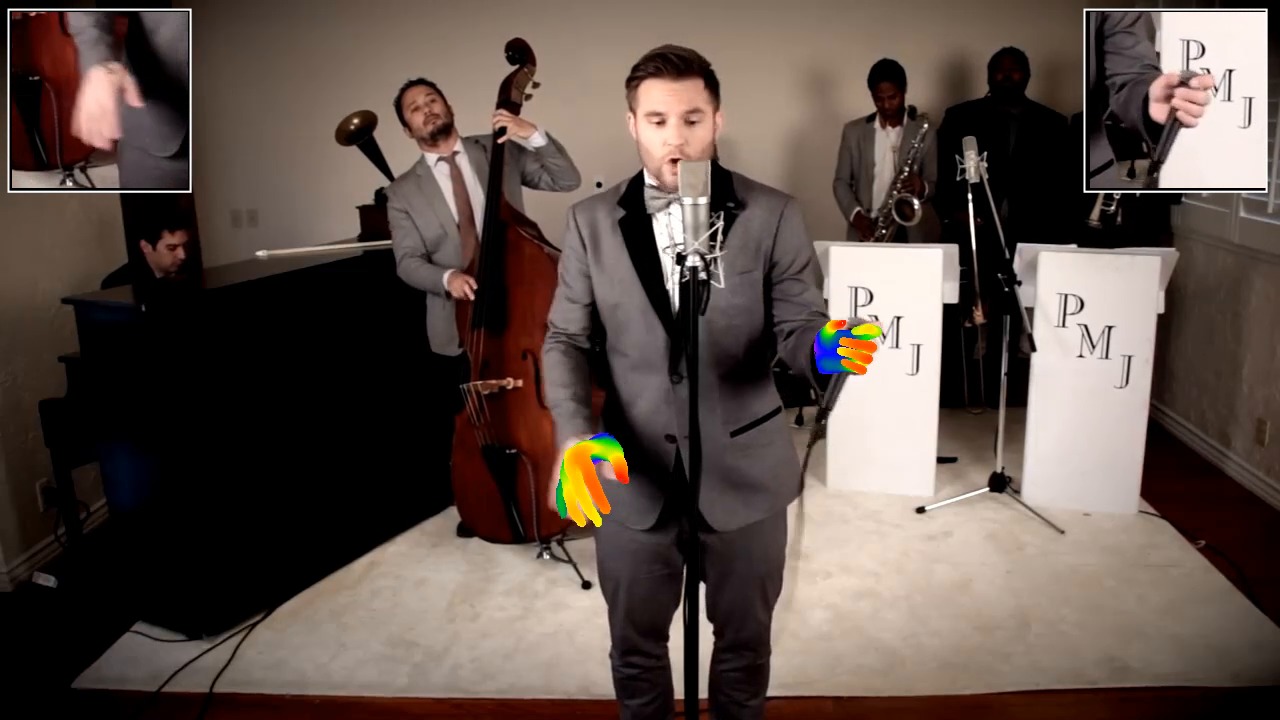}
\includegraphics[width=0.49\columnwidth]{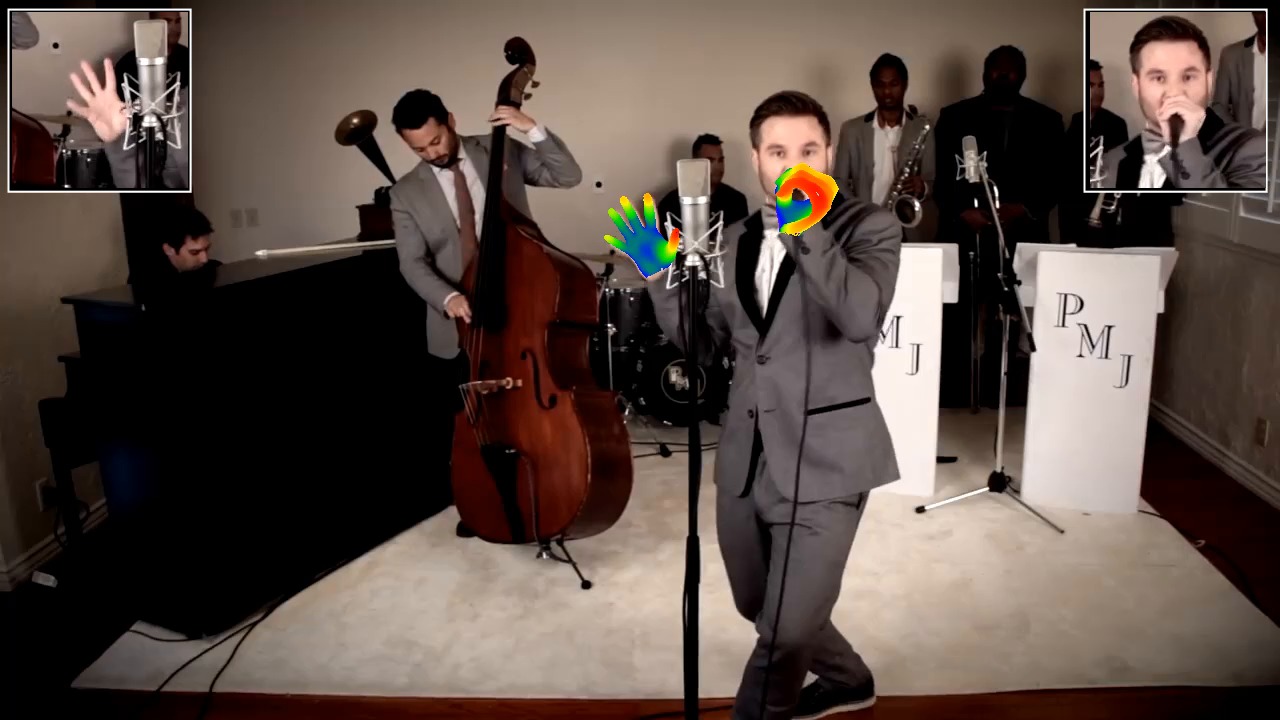}\\
\includegraphics[width=0.49\columnwidth]{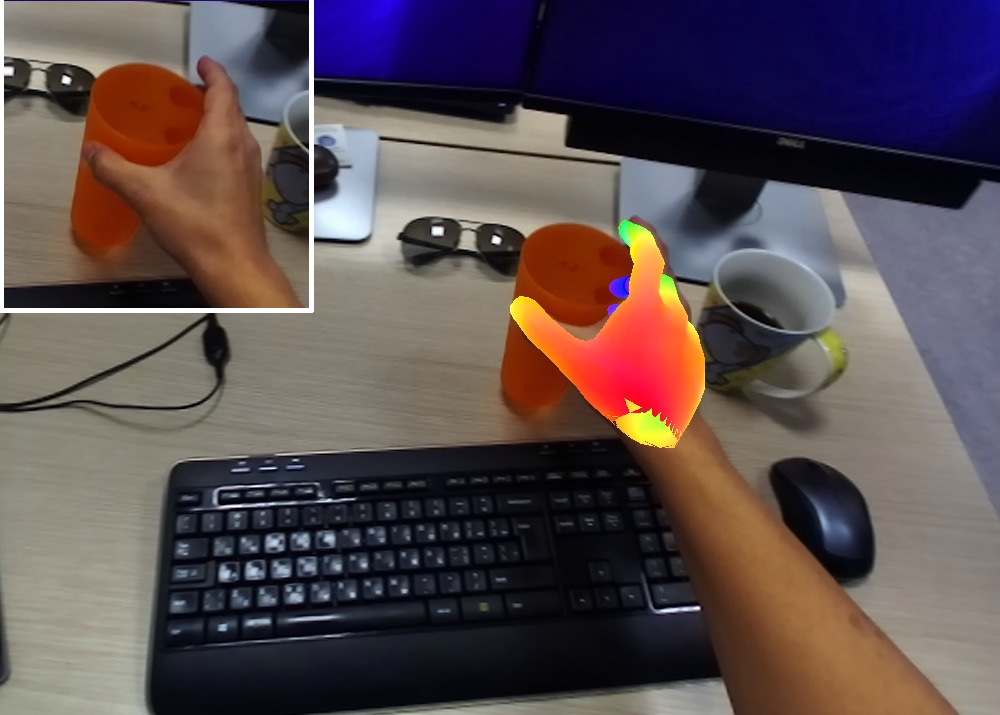}
\includegraphics[width=0.49\columnwidth]{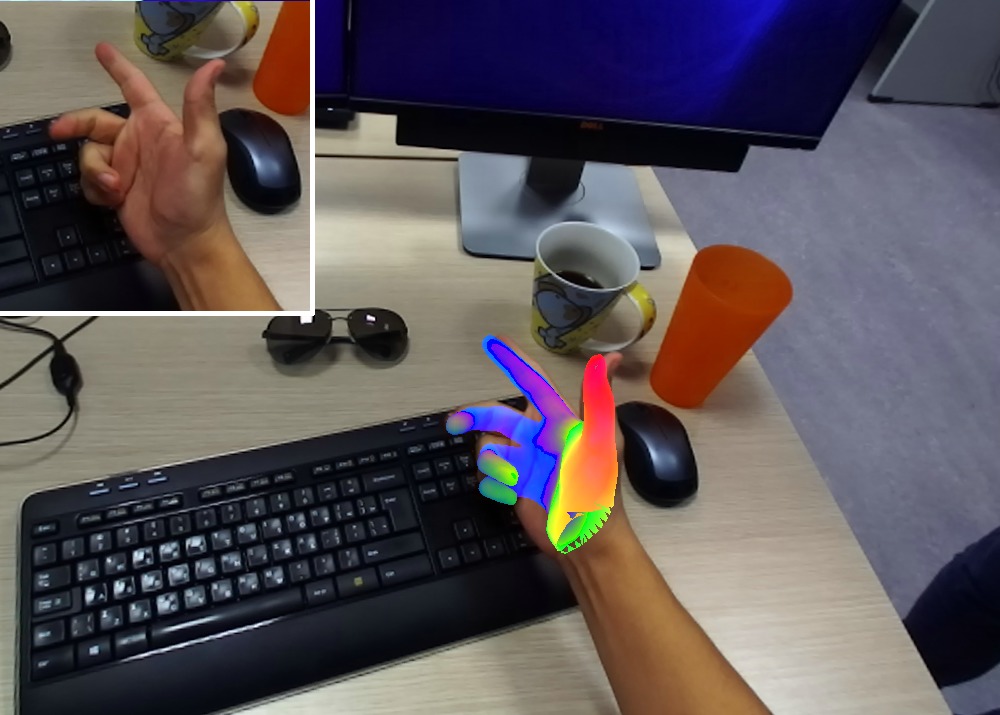}
\includegraphics[width=0.49\columnwidth]{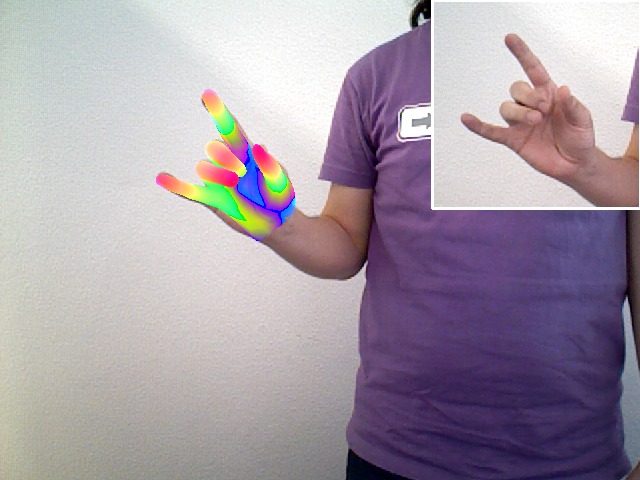}
\includegraphics[width=0.49\columnwidth]{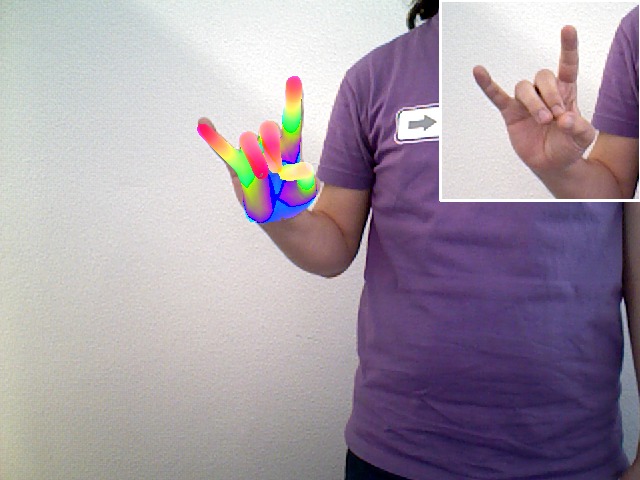}\\
\includegraphics[width=0.49\columnwidth]{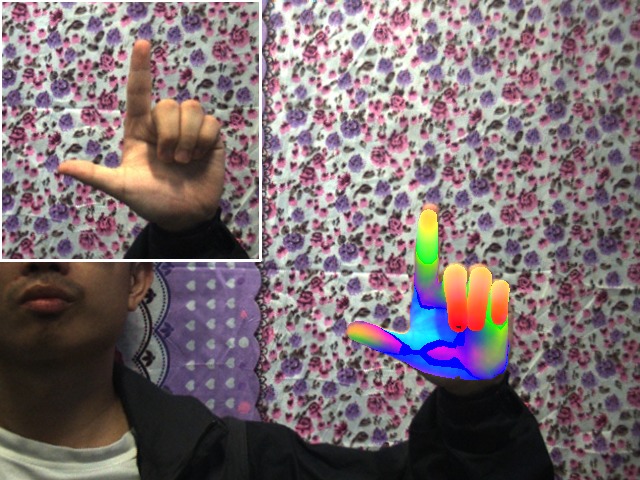}
\includegraphics[width=0.49\columnwidth]{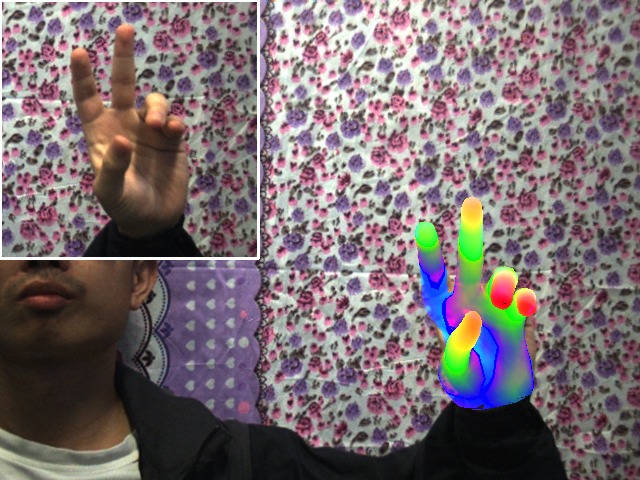}
\includegraphics[width=0.49\columnwidth]{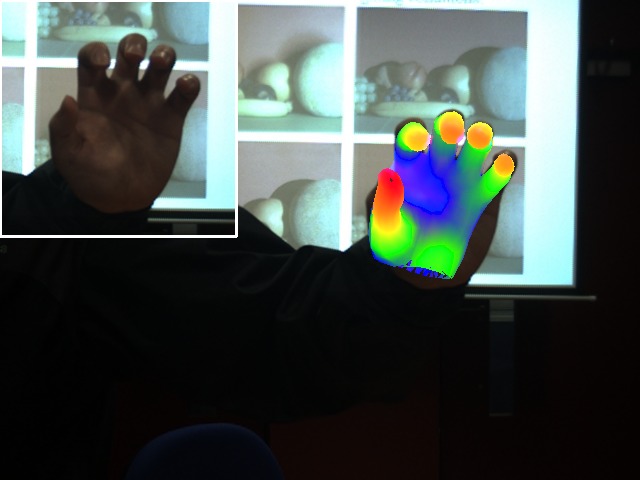}
\includegraphics[width=0.49\columnwidth]{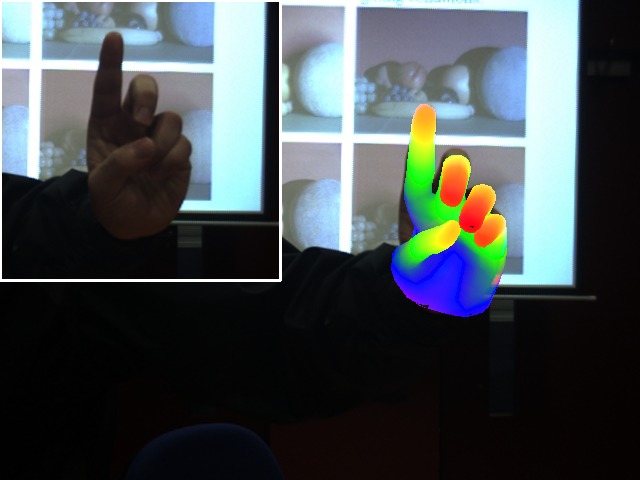}\\
\end{center}
    \caption{Sample qualitative 3D hand pose estimation results. Rows one, two and three: Frames from youtube videos. Complex hand poses with self-occlusions and object manipulation are showcased. Row four (left): Frames from an egocentric view. This sequence is also part of the supplementary video. Row four (right): Frames from the {\bf HIC} dataset. Row five: Frames from the {\bf SHD} dataset. In all images, the original hand(s) view is cropped and shown on the top corner.}
\label{fig:qual}
\end{figure*}  

\subsection{Qualitative results}
Figure~\ref{fig:qual} provides qualitative results from the application of the proposed 3D hand pose estimation method to a variety of image frames. These frames  belong to either standard datasets or have been extracted from youtube videos. In the case of dataset frames, the calibration provided by the authors was used. In the case of youtube videos, a generic calibration adjusted for the resolution of the input was selected. It can be verified that the estimated 3D hand models agree with the visual data even when generic calibration parameters were used. More indicative such results are included in the supplementary material accompanying the paper.  

The performed experiments (both the quantitative and the qualitative ones) also demonstrate that the hand model does not have to match exactly the observed one in order to estimate a 3D pose of reasonable accuracy.

\section{Summary}
We presented the first method that is capable of estimating the 3D pose of hands observed from monocular RGB input in the wild. The proposed hybrid approach consists of a discriminative component (used to estimate the location of 2D joints through a powerful CNN-based method) that is appropriately combined with the power of a generative approach (used to lift 2D joint estimations in 3D). We presented quantitative experimental evidence showing that our method compares favorably to the state of the art. Further qualitative experimental validation of the method was presented in challenging real-world situations. The presented solution avoids a series of limitations that hinder the deployment of 3D hand pose estimation in real life applications. Compared to tracking-based methods, initialization is not required. RGBD-based methods perform fairly accurately, nevertheless they have constraints regarding natural lighting and outdoor environments, spatial and temporal resolution and limitations on the effective range of observation. Stereo camera configurations have also been employed, nevertheless, they are more expensive and sensitive to calibration issues. On the contrary, RGB cameras are cheap and available everywhere. Thus, our approach can be useful in a broad spectrum of applications that require knowledge of the 3D pose of human hands.  

Future directions of research include the automatic adjustment of the model metrics to the observed hand, the incorporation of color cues in the last step of our approach and the investigation of approaches for automatic estimation of the camera calibration.

\subsection{Acknowledgments}
This work was partially supported by the projects Co4Robots (H2020-ICT-2016-1-731869) and WEARHAP (EUFP7-ICT-2011-9-601165).

{\small
\bibliographystyle{ieee}
\bibliography{monohand}

\begin{thebibliography}{10}\itemsep=-1pt

\bibitem{ceres}
S.~Agarwal, K.~Mierle, and Others.
\newblock Ceres solver.
\newblock \url{http://ceres-solver.org}.

\bibitem{mpii}
M.~Andriluka, L.~Pishchulin, P.~Gehler, and B.~Schiele.
\newblock 2d human pose estimation: New benchmark and state of the art
  analysis.
\newblock In {\em IEEE Conference on Computer Vision and Pattern Recognition
  (CVPR)}, June 2014.

\bibitem{andriluka2008people}
M.~Andriluka, S.~Roth, and B.~Schiele.
\newblock People-tracking-by-detection and people-detection-by-tracking.
\newblock In {\em Computer Vision and Pattern Recognition, 2008. CVPR 2008.
  IEEE Conference on}, pages 1--8. IEEE, 2008.

\bibitem{athitsos2003estimating}
V.~Athitsos and S.~Sclaroff.
\newblock Estimating 3d hand pose from a cluttered image.
\newblock In {\em CVPR}, 2003.

\bibitem{cao2017realtime}
Z.~Cao, T.~Simon, S.-E. Wei, and Y.~Sheikh.
\newblock Realtime multi-person 2d pose estimation using part affinity fields.
\newblock In {\em CVPR}, 2017.

\bibitem{chen2017pose}
X.~Chen, G.~Wang, H.~Guo, and C.~Zhang.
\newblock Pose guided structured region ensemble network for cascaded hand pose
  estimation.
\newblock {\em arXiv preprint arXiv:1708.03416}, 2017.

\bibitem{de2006regression}
T.~E. de~Campos and D.~W. Murray.
\newblock Regression-based hand pose estimation from multiple cameras.
\newblock In {\em CVPR}, 2006.

\bibitem{de2011model}
M.~de~La~Gorce, D.~J. Fleet, and N.~Paragios.
\newblock Model-based 3d hand pose estimation from monocular video.
\newblock {\em IEEE transactions on pattern analysis and machine intelligence},
  33(9):1793--1805, 2011.

\bibitem{erol2007vision}
A.~Erol, G.~Bebis, M.~Nicolescu, R.~D. Boyle, and X.~Twombly.
\newblock Vision-based hand pose estimation: A review.
\newblock {\em Computer Vision and Image Understanding}, 108(1):52--73, 2007.

\bibitem{ferrari2008progressive}
V.~Ferrari, M.~Marin-Jimenez, and A.~Zisserman.
\newblock Progressive search space reduction for human pose estimation.
\newblock In {\em Computer Vision and Pattern Recognition, 2008. CVPR 2008.
  IEEE Conference on}, pages 1--8. IEEE, 2008.

\bibitem{fleishman2015icpik}
S.~Fleishman, M.~Kliger, A.~Lerner, and G.~Kutliroff.
\newblock Icpik: Inverse kinematics based articulated-icp.
\newblock In {\em CVPRW}, 2015.

\bibitem{ge2016robust}
L.~Ge, H.~Liang, J.~Yuan, and D.~Thalmann.
\newblock Robust 3d hand pose estimation in single depth images: from
  single-view cnn to multi-view cnns.
\newblock In {\em Proceedings of the IEEE Conference on Computer Vision and
  Pattern Recognition}, pages 3593--3601, 2016.

\bibitem{ge20173d}
L.~Ge, H.~Liang, J.~Yuan, and D.~Thalmann.
\newblock 3d convolutional neural networks for efficient and robust hand pose
  estimation from single depth images.
\newblock In {\em Proceedings of the IEEE Conference on Computer Vision and
  Pattern Recognition}, pages 1991--2000, 2017.

\bibitem{girshick2011efficient}
R.~Girshick, J.~Shotton, P.~Kohli, A.~Criminisi, and A.~Fitzgibbon.
\newblock Efficient regression of general-activity human poses from depth
  images.
\newblock In {\em Computer Vision (ICCV), 2011 IEEE International Conference
  on}, pages 415--422. IEEE, 2011.

\bibitem{gomez2017large}
F.~Gomez-Donoso, S.~Orts-Escolano, and M.~Cazorla.
\newblock Large-scale multiview 3d hand pose dataset.
\newblock {\em arXiv preprint arXiv:1707.03742}, 2017.

\bibitem{guo2017towards}
H.~Guo, G.~Wang, X.~Chen, and C.~Zhang.
\newblock Towards good practices for deep 3d hand pose estimation.
\newblock {\em arXiv preprint arXiv:1707.07248}, 2017.

\bibitem{keskin2012hand}
C.~Keskin, F.~K{\i}ra{\c{c}}, Y.~E. Kara, and L.~Akarun.
\newblock Hand pose estimation and hand shape classification using
  multi-layered randomized decision forests.
\newblock In {\em ECCV}, 2012.

\bibitem{khamis2015learning}
S.~Khamis, J.~Taylor, J.~Shotton, C.~Keskin, S.~Izadi, and A.~Fitzgibbon.
\newblock Learning an efficient model of hand shape variation from depth
  images.
\newblock In {\em CVPR}, 2015.

\bibitem{kyriazis2013physically}
N.~Kyriazis and A.~Argyros.
\newblock Physically plausible 3d scene tracking: The single actor hypothesis.
\newblock In {\em CVPR}, 2013.

\bibitem{li20153d}
P.~Li, H.~Ling, X.~Li, and C.~Liao.
\newblock 3d hand pose estimation using randomized decision forest with
  segmentation index points.
\newblock In {\em ICCV}, 2015.

\bibitem{mscoco}
T.-Y. Lin, M.~Maire, S.~Belongie, J.~Hays, P.~Perona, D.~Ramanan,
  P.~Doll{\'a}r, and C.~L. Zitnick.
\newblock {\em Microsoft COCO: Common Objects in Context}, pages 740--755.
\newblock Springer International Publishing, Cham, 2014.

\bibitem{liu2016ssd}
W.~Liu, D.~Anguelov, D.~Erhan, C.~Szegedy, S.~Reed, C.-Y. Fu, and A.~C. Berg.
\newblock Ssd: Single shot multibox detector.
\newblock In {\em European conference on computer vision}, pages 21--37.
  Springer, 2016.

\bibitem{MakrisArgyros2015a}
A.~Makris and A.~A. Argyros.
\newblock Model-based 3d hand tracking with on-line shape adaptation.
\newblock In {\em BMVC}, 2015.

\bibitem{MakrisKyriazisArgyros2015a}
A.~Makris, N.~Kyriazis, and A.~A. Argyros.
\newblock Hierarchical particle filtering for 3d hand tracking.
\newblock In {\em CVPR}, 2015.

\bibitem{mehta2017vnect}
D.~Mehta, S.~Sridhar, O.~Sotnychenko, H.~Rhodin, M.~Shafiei, H.-P. Seidel,
  W.~Xu, D.~Casas, and C.~Theobalt.
\newblock Vnect: Real-time 3d human pose estimation with a single rgb camera.
\newblock In {\em ACM Transactions on Graphics}, volume~36, July 2017.

\bibitem{melax2013dynamics}
S.~Melax, L.~Keselman, and S.~Orsten.
\newblock Dynamics based 3d skeletal hand tracking.
\newblock In {\em Proc. of Graphics Interface}, 2013.

\bibitem{moreno20173d}
F.~Moreno-Noguer.
\newblock 3d human pose estimation from a single image via distance matrix
  regression.
\newblock In {\em 2017 IEEE Conference on Computer Vision and Pattern
  Recognition (CVPR)}, pages 1561--1570. IEEE, 2017.

\bibitem{oberweger2017deeppriorplus}
M.~Oberweger and V.~Lepetit.
\newblock Deepprior++: Improving fast and accurate 3d hand pose estimation.
\newblock In {\em ICCV workshop}, volume 840, 2017.

\bibitem{oberweger2016efficiently}
M.~Oberweger, G.~Riegler, P.~Wohlhart, and V.~Lepetit.
\newblock Efficiently creating 3d training data for fine hand pose estimation.
\newblock In {\em Proceedings of the IEEE Conference on Computer Vision and
  Pattern Recognition}, pages 4957--4965, 2016.

\bibitem{oberweger2015hands}
M.~Oberweger, P.~Wohlhart, and V.~Lepetit.
\newblock Hands deep in deep learning for hand pose estimation.
\newblock In {\em Computer Vision Winter Workshop}, pages 21--30, 2015.

\bibitem{oberweger2015training}
M.~Oberweger, P.~Wohlhart, and V.~Lepetit.
\newblock Training a feedback loop for hand pose estimation.
\newblock In {\em ICCV}, 2015.

\bibitem{oikonomidis2010markerless}
I.~Oikonomidis, N.~Kyriazis, and A.~A. Argyros.
\newblock Markerless and efficient 26-dof hand pose recovery.
\newblock In {\em ACCV}, 2010.

\bibitem{oikonomidis2011efficient}
I.~Oikonomidis, N.~Kyriazis, and A.~A. Argyros.
\newblock Efficient model-based 3d tracking of hand articulations using kinect.
\newblock In {\em BMVC}, 2011.

\bibitem{panteleris2017back}
P.~Panteleris and A.~Argyros.
\newblock Back to rgb: 3d tracking of hands and hand-object interactions based
  on short-baseline stereo.
\newblock {\em arXiv preprint arXiv:1705.05301}, 2017.

\bibitem{poier2015hybrid}
G.~Poier, K.~Roditakis, S.~Schulter, D.~Michel, H.~Bischof, and A.~A. Argyros.
\newblock Hybrid one-shot 3d hand pose estimation by exploiting uncertainties.
\newblock In {\em British Machine Vision Conference (BMVC 2015)}, pages 182--1,
  Swansea, UK, September 2015. BMVA.

\bibitem{qian2014realtime}
C.~Qian, X.~Sun, Y.~Wei, X.~Tang, and J.~Sun.
\newblock Realtime and robust hand tracking from depth.
\newblock In {\em CVPR}, 2014.

\bibitem{redmon2016you}
J.~Redmon, S.~Divvala, R.~Girshick, and A.~Farhadi.
\newblock You only look once: Unified, real-time object detection.
\newblock In {\em Proceedings of the IEEE Conference on Computer Vision and
  Pattern Recognition}, pages 779--788, 2016.

\bibitem{redmon2017yolo9000}
J.~Redmon and A.~Farhadi.
\newblock Yolo9000: Better, faster, stronger.
\newblock In {\em Computer Vision and Pattern Recognition, 2017. CVPR 2017.
  IEEE Conference on}, 2017.

\bibitem{rehg1994visual}
J.~M. Rehg and T.~Kanade.
\newblock Visual tracking of high dof articulated structures: an application to
  human hand tracking.
\newblock In {\em ECCV}, 1994.

\bibitem{ren2017faster}
S.~Ren, K.~He, R.~Girshick, and J.~Sun.
\newblock Faster r-cnn: Towards real-time object detection with region proposal
  networks.
\newblock {\em IEEE transactions on pattern analysis and machine intelligence},
  39(6):1137--1149, 2017.

\bibitem{rogez2015first}
G.~Rogez, J.~S. Supancic, and D.~Ramanan.
\newblock First-person pose recognition using egocentric workspaces.
\newblock In {\em CVPR}, 2015.

\bibitem{romero2009monocular}
J.~Romero, H.~Kjellstr{\"o}m, and D.~Kragic.
\newblock Monocular real-time 3d articulated hand pose estimation.
\newblock In {\em Humanoid Robots, 2009. Humanoids 2009. 9th IEEE-RAS
  International Conference on}, pages 87--92. IEEE, 2009.

\bibitem{romero2010hands}
J.~Romero, H.~Kjellstr{\"o}m, and D.~Kragic.
\newblock Hands in action: real-time 3d reconstruction of hands in interaction
  with objects.
\newblock In {\em ICRA}, 2010.

\bibitem{sharp2015accurate}
T.~Sharp, C.~Keskin, D.~Robertson, J.~Taylor, J.~Shotton, D.~Kim, C.~Rhemann,
  I.~Leichter, A.~Vinnikov, Y.~Wei, et~al.
\newblock Accurate, robust, and flexible real-time hand tracking.
\newblock In {\em Proceedings of the 33rd Annual ACM Conference on Human
  Factors in Computing Systems}, 2015.

\bibitem{shotton2013real}
J.~Shotton, T.~Sharp, A.~Kipman, A.~Fitzgibbon, M.~Finocchio, A.~Blake,
  M.~Cook, and R.~Moore.
\newblock Real-time human pose recognition in parts from single depth images.
\newblock {\em Communications of the ACM}, 56(1):116--124, 2013.

\bibitem{simon2017hand}
T.~Simon, H.~Joo, I.~Matthews, and Y.~Sheikh.
\newblock Hand keypoint detection in single images using multiview
  bootstrapping.
\newblock In {\em CVPR}, 2017.

\bibitem{simonyan2014very}
K.~Simonyan and A.~Zisserman.
\newblock Very deep convolutional networks for large-scale image recognition.
\newblock {\em arXiv preprint arXiv:1409.1556}, 2014.

\bibitem{sinha2016deephand}
A.~Sinha, C.~Choi, and K.~Ramani.
\newblock Deephand: Robust hand pose estimation by completing a matrix imputed
  with deep features.
\newblock In {\em Proceedings of the IEEE Conference on Computer Vision and
  Pattern Recognition}, pages 4150--4158, 2016.

\bibitem{sridhar2015fast}
S.~Sridhar, F.~Mueller, A.~Oulasvirta, and C.~Theobalt.
\newblock Fast and robust hand tracking using detection-guided optimization.
\newblock In {\em CVPR}, 2015.

\bibitem{sridhar2013interactive}
S.~Sridhar, A.~Oulasvirta, and C.~Theobalt.
\newblock Interactive markerless articulated hand motion tracking using rgb and
  depth data.
\newblock In {\em ICCV}, 2013.

\bibitem{sridhar2014real}
S.~Sridhar, H.~Rhodin, H.-P. Seidel, A.~Oulasvirta, and C.~Theobalt.
\newblock Real-time hand tracking using a sum of anisotropic gaussians model.
\newblock In {\em 3DV}, 2014.

\bibitem{stenger2001model}
B.~Stenger, P.~R. Mendon{\c{c}}a, and R.~Cipolla.
\newblock Model-based 3d tracking of an articulated hand.
\newblock In {\em CVPR}, 2001.

\bibitem{sudderth2004visual}
E.~B. Sudderth, M.~I. Mandel, W.~T. Freeman, and A.~S. Willsky.
\newblock Visual hand tracking using nonparametric belief propagation.
\newblock In {\em CVPRW}, 2004.

\bibitem{sun2017compositional}
X.~Sun, J.~Shang, S.~Liang, and Y.~Wei.
\newblock Compositional human pose regression.
\newblock In {\em The IEEE International Conference on Computer Vision (ICCV)},
  Oct 2017.

\bibitem{sun2015cascaded}
X.~Sun, Y.~Wei, S.~Liang, X.~Tang, and J.~Sun.
\newblock Cascaded hand pose regression.
\newblock In {\em CVPR}, 2015.

\bibitem{sung2012unstructured}
J.~Sung, C.~Ponce, B.~Selman, and A.~Saxena.
\newblock Unstructured human activity detection from rgbd images.
\newblock In {\em Robotics and Automation (ICRA), 2012 IEEE International
  Conference on}, pages 842--849. IEEE, 2012.

\bibitem{tagliasacchi2015robust}
A.~Tagliasacchi, M.~Schr{\"o}der, A.~Tkach, S.~Bouaziz, M.~Botsch, and
  M.~Pauly.
\newblock Robust articulated-icp for real-time hand tracking.
\newblock In {\em Computer Graphics Forum}, 2015.

\bibitem{tang2014latent}
D.~Tang, H.~Jin~Chang, A.~Tejani, and T.-K. Kim.
\newblock Latent regression forest: Structured estimation of 3d articulated
  hand posture.
\newblock In {\em Proceedings of the IEEE Conference on Computer Vision and
  Pattern Recognition}, pages 3786--3793, 2014.

\bibitem{tang2015opening}
D.~Tang, J.~Taylor, P.~Kohli, C.~Keskin, T.-K. Kim, and J.~Shotton.
\newblock Opening the black box: Hierarchical sampling optimization for
  estimating human hand pose.
\newblock In {\em ICCV}, 2015.

\bibitem{tang2013real}
D.~Tang, T.-H. Yu, and T.-K. Kim.
\newblock Real-time articulated hand pose estimation using semi-supervised
  transductive regression forests.
\newblock In {\em ICCV}, 2013.

\bibitem{taylor2016efficient}
J.~Taylor, L.~Bordeaux, T.~Cashman, B.~Corish, C.~Keskin, T.~Sharp, E.~Soto,
  D.~Sweeney, J.~Valentin, B.~Luff, et~al.
\newblock Efficient and precise interactive hand tracking through joint,
  continuous optimization of pose and correspondences.
\newblock {\em ACM TOG}, 2016.

\bibitem{taylor2014user}
J.~Taylor, R.~Stebbing, V.~Ramakrishna, C.~Keskin, J.~Shotton, S.~Izadi,
  A.~Hertzmann, and A.~Fitzgibbon.
\newblock User-specific hand modeling from monocular depth sequences.
\newblock In {\em CVPR}, 2014.

\bibitem{tompson2014real}
J.~Tompson, M.~Stein, Y.~Lecun, and K.~Perlin.
\newblock Real-time continuous pose recovery of human hands using convolutional
  networks.
\newblock {\em ACM TOG}, 2014.

\bibitem{tompson2014joint}
J.~J. Tompson, A.~Jain, Y.~LeCun, and C.~Bregler.
\newblock Joint training of a convolutional network and a graphical model for
  human pose estimation.
\newblock In {\em Advances in neural information processing systems}, pages
  1799--1807, 2014.

\bibitem{toshev2014deeppose}
A.~Toshev and C.~Szegedy.
\newblock Deeppose: Human pose estimation via deep neural networks.
\newblock In {\em Proceedings of the IEEE Conference on Computer Vision and
  Pattern Recognition}, pages 1653--1660, 2014.

\bibitem{tzionas2015capturing}
D.~Tzionas, L.~Ballan, A.~Srikantha, P.~Aponte, M.~Pollefeys, and J.~Gall.
\newblock Capturing hands in action using discriminative salient points and
  physics simulation.
\newblock {\em IJCV}, 2015.

\bibitem{wan2016hand}
C.~Wan, A.~Yao, and L.~Van~Gool.
\newblock Hand pose estimation from local surface normals.
\newblock In {\em European Conference on Computer Vision}, pages 554--569.
  Springer, 2016.

\bibitem{wei2016}
S.-E. Wei, V.~Ramakrishna, T.~Kanade, and Y.~Sheikh.
\newblock Convolutional pose machines.
\newblock In {\em The IEEE Conference on Computer Vision and Pattern
  Recognition (CVPR)}, June 2016.

\bibitem{xu2013efficient}
C.~Xu and L.~Cheng.
\newblock Efficient hand pose estimation from a single depth image.
\newblock In {\em ICCV}, 2013.

\bibitem{zhang20163d}
J.~Zhang, J.~Jiao, M.~Chen, L.~Qu, X.~Xu, and Q.~Yang.
\newblock 3d hand pose tracking and estimation using stereo matching.
\newblock {\em arXiv:1610.07214}, 2016.

\bibitem{zhou2016model}
X.~Zhou, Q.~Wan, W.~Zhang, X.~Xue, and Y.~Wei.
\newblock Model-based deep hand pose estimation.
\newblock In {\em IJCAI}, 2016.

\bibitem{zimmermann2017learning}
C.~Zimmermann and T.~Brox.
\newblock Learning to estimate 3d hand pose from single rgb images.
\newblock In {\em The IEEE International Conference on Computer Vision (ICCV)},
  Oct 2017.

\end{thebibliography}
}

\end{document}